\documentclass[runningheads]{llncs}

\usepackage{eccv}

\usepackage{eccvabbrv}

\usepackage{graphicx}
\usepackage{booktabs}

\usepackage[accsupp]{axessibility}  %

\usepackage{xcolor}
\usepackage{multicol,multirow}
\usepackage{overpic}
\usepackage{float}
\usepackage{wrapfig}
\usepackage{amsmath}
\usepackage{amssymb}
\usepackage{afterpage}
\usepackage{pifont}%
\usepackage{wrapfig}
\usepackage{makecell}
\usepackage{dsfont}
\usepackage{nccmath}
\usepackage{mathtools}
\usepackage{nicefrac}
\usepackage{listings}
\usepackage{rotating}
\usepackage{enumitem}

\usepackage[dvipsnames]{xcolor}

\newcommand{\methodname}[0]{ECHO}
\newcommand{\mat}[1]{\boldsymbol{#1}}
\renewcommand{\vec}[1]{\boldsymbol{#1}}
\newcommand{\set}[1]{\mathcal{#1}}
\newcommand{\real}[0]{\mathbb{R}}
\newcommand{\window}[0]{W}
\newcommand{\sequence}[0]{N}
\newcommand{\methodparams}[0]{\psi}
\newcommand{\methodnetwork}[0]{\mathrm{ECHO}_\psi}
\definecolor{cond_col}{RGB}{204, 204, 0}
\definecolor{gt_col}{RGB}{200, 220, 245}
\definecolor{noise_col}{RGB}{20, 20, 100}
\definecolor{t_col}{RGB}{154,1,150}

\newcommand{\h}[0]{\set{H} }
\newcommand{\hglobal}[0]{\mat{T}_\h }
\newcommand{\hshape}[0]{\vec{\beta}}
\newcommand{\hface}[0]{\vec{\psi}}
\newcommand{\hposefull}[0]{\vec{\theta} }
\newcommand{\hpose}[0]{\vec{\theta}_\h }

\newcommand{\htemplatev}[0]{\mat{V}_\h }
\newcommand{\thuman}[0]{\set{T}_\h }
\newcommand{\hjoints}[0]{\mat{J}_\h }
\newcommand{\hvelocity}[0]{\mat{U}_\h }

\newcommand{\obj}[0]{\set{O} }
\newcommand{\objglobal}[0]{\mat{T_\obj} }
\newcommand{\objonehot}[0]{\vec{y}_\obj }
\newcommand{\objfeatures}[0]{\vec{f}_\obj }
\newcommand{\objtemplatev}[0]{\mat{V}_\obj }
\newcommand{\condobj}[0]{\set{C}_{\obj} }
\newcommand{\tobj}[0]{\set{T}_\obj }

\newcommand{\inter}[0]{\set{I} }
\newcommand{\intervec}[0]{\vec{c}_{\inter}}
\newcommand{\interpoints}[0]{\mat{P}_{c}}
\newcommand{\tinter}[0]{\set{T}_\inter }

\newcommand{\condego}[0]{\set{E} }
\newcommand{\egoheaddeltat}[0]{\Delta \mat{T}^{t-1, t}_{\text{head}} }
\newcommand{\egohanddeltat}[0]{\Delta \mat{T}^{t-1, t}_{\text{hands}} }
\newcommand{\egorotcan}[0]{\mat{R}^{t}_{\text{can,head}} }
\newcommand{\egorothandscan}[0]{\mat{R}^{t}_{\text{can,hands}} }
\newcommand{\egoheight}[0]{ h^{t}_{\text{head}} }

\newcommand{\predhpose}[0]{\widehat{\vec{\theta}}_\h }

\newcommand{\predhjoints}[0]{\widehat{\mat{J}}_\h }
\newcommand{\predhvelocity}[0]{\widehat{\vec{U}}_\h }
\newcommand{\predobjvelocity}[0]{\widehat{\vec{U}}_\obj }
\newcommand{\predobjangvelocity}[0]{\widehat{\vec{\omega}}_\obj }
\newcommand{\predobjglobal}[0]{\widehat{\mat{T}}_\obj }
\newcommand{\predobjtemplatev}[0]{\widehat{\mat{V}}_\obj }
\newcommand{\predhtemplatev}[0]{\widehat{\mat{V}}_\h }
\newcommand{\predintervec}[0]{\widehat{\vec{c}}_\inter }

\newcommand{\bodiffo}[1][]{BoDiffusion#1+$\mathsf{O}$}
\newcommand{\egoalloho}[1][]{EgoAllo#1+$\mathsf{H}$+$\mathsf{O}$}
\newcommand{\first}[1]{\mat{#1}}
\newcommand{\second}[1]{\underline{#1}}
\newcommand{\rottext}[2][90]{%
    \rotatebox[origin=c]{#1}{%
        \parbox[c]{2cm}{
            \centering\arraybackslash #2%
        }%
    }%
}

\usepackage[pagebackref,breaklinks,colorlinks,citecolor=eccvblue]{hyperref}

\usepackage{orcidlink}

\begin{document}

\title{ECHO: Ego-Centric modeling of Human-Object interactions} 

\titlerunning{ECHO: Ego-Centric modeling of Human-Object interactions}

\author{Ilya A. Petrov\inst{1,2}\orcidlink{0000-0002-8900-1071} \and
Vladimir Guzov\inst{1,2}\orcidlink{0000-0003-1304-5577} \and
Riccardo Marin\inst{3,4}\orcidlink{0000-0003-2392-4612} \and
Emre Aksan\inst{5}\orcidlink{0000-0002-9836-9011} \and 
Xu Chen\inst{5} \and 
Daniel Cremers\inst{3,4}\orcidlink{0000-0002-3079-7984} \and 
Thabo Beeler\inst{5}\orcidlink{0000-0002-8077-1205} \and 
Gerard Pons-Moll\inst{1,2,6}\orcidlink{0000-0001-5115-7794}}

\authorrunning{I.A.~Petrov et al.}

\institute{
$^{1}${\small University of T\"ubingen, Germany}\qquad
$^{2}$ {\small T\"ubingen AI Center, Germany}\\
$^{3}$ {\small Technical University of Munich, Germany}\\
$^{4}$ {\small Munich Center for Machine Learning, Germany}\qquad
$^{5}$ {\small Google, Switzerland}\\
$^{6}$ {\small Max Planck Institute for Informatics, Saarland Informatics Campus, Germany}\\
\url{https://virtualhumans.mpi-inf.mpg.de/echo/}}

\maketitle

\begin{abstract}
Modeling human-object interactions (HOI) from an egocentric perspective is a critical yet challenging task, particularly when relying on sparse signals from wearable devices like smart glasses and watches. 
We present {\methodname}, the first unified framework to jointly recover human pose, object motion, and contact dynamics solely from head and wrist tracking. 
To tackle the underconstrained nature of this problem, we introduce a novel tri-variate diffusion process with independent noise schedules that models the mutual dependencies between the human, object, and interaction modalities. 
This formulation allows {\methodname} to operate with flexible input configurations, making it robust to intermittent tracking and capable of leveraging partial observations. 
Crucially, it enables training on a combination of large-scale human motion datasets and smaller HOI collections, learning strong priors while capturing interaction nuances.
Furthermore, we employ a smooth inpainting inference mechanism that enables the generation of temporally consistent interactions for arbitrarily long sequences. 
Extensive evaluations demonstrate that {\methodname} achieves state-of-the-art performance, significantly outperforming existing methods lacking such flexibility.
\end{abstract}

\section{Introduction}
\begin{figure}[ht!]
    \centering
    \includegraphics[trim=0cm 0cm 0cm 0cm,clip,width=1.0\linewidth]{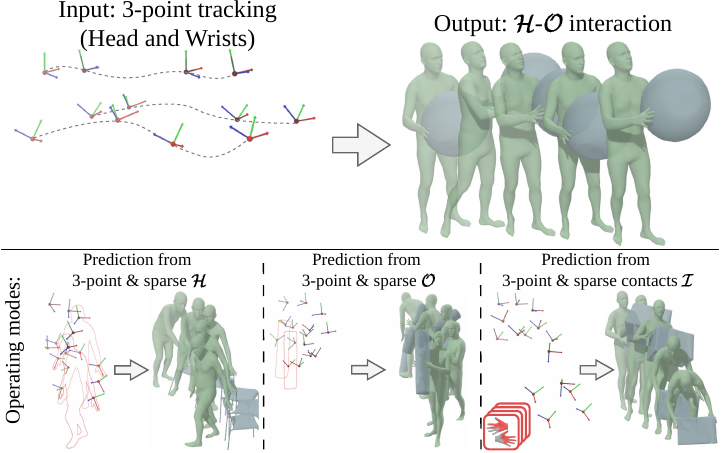}
    \vspace{-8pt}
    \caption{\textbf{{\methodname}}. Inferring complex interactions from sparse wearable signals is challenging. {\methodname} is the first method to jointly recover full-body Human-Object Interaction sequences (top) solely from sparse 3-point tracking. Our flexible framework supports various inference modes (bottom), leveraging partial or intermittent observations (shown in {\color{red}{red}}) of human pose, object trajectory, or contact dynamics.}
    \label{fig:teaser}
    \vspace{-10pt}
\end{figure}
\label{sec:introduction}
Wearable sensors like smart glasses~\cite{Aria}, rings~\cite{OpenRing}, and wristbands~\cite{MetaEMG} are becoming ubiquitous.
Beyond enabling XR experiences, they serve as everyday companions that continuously monitor user activities. 
Consequently, developing algorithms to robustly perceive human-object interactions from such sparse signals is a key research challenge with far-reaching impact, unlocking applications in healthcare, personal assistants, entertainment, robotics, and spatial AI.

Recent work~\cite{guzov-jiang2025hmd2} demonstrates that it is possible to reconstruct users' motion in-the-wild from sparse sensors.
Despite these advances, egocentric human-object interaction (HOI) remains underexplored.
Existing methods~\cite{guzov24ireplica, yang2024egochoir} typically require RGB images, pre-scanned scenes, or specialized capturing suits, and rely on hand-crafted constraints, restricting scalability and generalization.
While learning-based approaches offer a promising alternative, current HOI datasets remain limited in scale and diversity, especially when compared to large human motion collections such as AMASS~\cite{AMASS:ICCV:2019}.
Hence, what is missing is a unified egocentric HOI method capable of learning from multiple modalities jointly.

We address this gap with {\methodname} (Fig.~\ref{fig:teaser}), the first method to model full-body HOI from head and hand tracking over arbitrary-length sequences. 
{\methodname} simultaneously predicts human pose, object trajectory, and contact dynamics.
While anchored in 3-point tracking, it can additionally leverage \textit{sparse} observations from any of these modalities (e.g., partial object tracking) to further constrain its predictions.
The model is also robust to intermittent hand tracking, which is common in real-world scenarios, and has no constraints on the sequence length.

We achieve this through a unified framework with several innovations.
The core of the method is a tri-variate diffusion formulation that effectively learns the inter-modal relationships between human, object, and contacts.
This design enables training on a combination of datasets, including human-only motion collections such as AMASS~\cite{AMASS:ICCV:2019} and HOI datasets like BEHAVE~\cite{bhatnagar2022behave} and OMOMO~\cite{li2023object}.
By using these data sources within one generative framework, the model learns a strong human motion prior while simultaneously learning interaction-specific dynamics, something existing HOI modeling methods cannot achieve.

Crucially, our approach supports flexible conditioning, leveraging partial observations of any modality to ensure robustness against sensor noise and intermittent tracking.
By explicitly modeling both human-environment and human-object contacts, we further enable self-supervised inference guidance to ensure physically plausible interactions.
Finally, we extend the inference approach of HMD\textsuperscript{2}~\cite{guzov-jiang2025hmd2} with a novel smooth inpainting that blends past and current predictions, enabling seamless, real-time processing of arbitrarily long sequences.

{\methodname} relies on minimal assumptions, making it flexible and adaptable to diverse settings.
Its tri-variate diffusion formulation naturally accommodates additional modalities (e.g., human tracking from IMUs or object tracking from RGB), making {\methodname} a universal approach for egocentric HOI modeling.
{\methodname} achieves state-of-the-art performance, surpassing competitors that lack the same flexibility.
Through detailed evaluation, we demonstrate the effectiveness of our design choices, which we believe will be instrumental for further research in egocentric perception and interaction modeling.

Our key contributions are:
\begin{itemize}[topsep=0pt]
    \item We present the first method to reconstruct HOI from wearable 3-point tracking, jointly recovering human motion, object trajectory, and contact.
    \item We introduce a unified tri-variate diffusion model that supports flexible cross-modal conditioning and partial observations, is robust to sensor noise, and performs inference for arbitrary-length sequences.
    \item We achieve state-of-the-art performance and validate our design through extensive ablations. The code and trained model are available at \url{https://github.com/ptrvilya/echo}.
\end{itemize}

\section{Related Work}
\label{sec:related_work}
\subsection{Egocentric Motion Reconstruction}
Using body-worn and head-mounted sensors for human motion reconstruction is an emerging research area.
Early methods used body-worn cameras to recover hand positions~\cite{ijcai2017-200,fathi2011understanding,ma2016going,cao2017egocentric,yonemoto2015egocentric,rogez2015first} or full-body motion~\cite{xu2019mo2cap2,tome2020selfpose,rhodin2016egocap,akada2022unrealego,kang2023ego3dpose,liu2023egofish3d,liu2023egohmr,lee2025rewind}.
These approaches focus on joint angles, ignoring the user-scene relation.

Other works utilize body-worn sensors like EMs~\cite{kaufmann2021pose}, EMGs~\cite{chiquier2023muscles}, and most popularly, IMUs~\cite{vlasic2007practical,von2017sparse,yi2021transpose,yi2022physical,DIP:SIGGRAPHAsia:2018,mollyn2023imuposer,jiang2022transformer,zuo2024loose,xu2024mobileposer}; however, double-integrating acceleration for position induces drift.

HPS~\cite{HPS} advanced the field by fusing IMU tracking with camera localization for global pose estimation in large scenes. Follow-up works enabled scene scanning~\cite{dai2022hsc4d,liu2024egohdm}, scaled datasets~\cite{ma2024nymeria, zhang2022egobody, hollidt2024egosim, de2025monado, yoon2026egoxtreme}, reduced sensor counts~\cite{jiang2022avatarposer, zheng2023realistic, winkler2022questsim,lee2024mocap,zheng2023realistic,dai2024hmd,jiang2024egoposer}, conditioned on past observations~\cite{barquero2025sparse}, and integrated biomechanical constraints~\cite{jiang2024manikin}.
Generative diffusion models enable realistic motion synthesis from underconstrained inputs~\cite{du2023avatars,castillo2023bodiffusion,wang2024egocentric}. 
For instance, EgoEgo~\cite{li2023ego} generates full-body motion from head trajectories, while LookOut~\cite{pan2025lookout} recovers head trajectories from in-the-wild video.

Recent works~\cite{guzov-jiang2025hmd2,yi2025egoallo,escobar2025egocast,chi2024estimating,xia2025envposer,patel2025uniegomotion, shin2026egomdm} utilize modalities like RGB, point clouds, and hand detections, exploring appearance modeling~\cite{feng2024stratified} and multi-modal fusion~\cite{hong2025egolm}.

We similarly use sparse head and hand conditioning. 
However, unlike these methods, our approach models dynamic interactions with objects, enabling a more complete reconstruction of human activity in real-world settings.

\subsection{Exocentric HOI Modeling}
Early HOI reconstruction focused on human-centric modeling in static environments~\cite{mir2024generating, hassan2021populating, huang2023diffusion, zhang2024scenic, hassan2019resolving,xu2024regennet,tang2024unified}, including recent work on reconstructing and placing humans within scenes~\cite{ym2026graft, ym2025physic, kister2026inhabit}.
These methods do not model object displacement in the scenes, limiting their applicability in real-world scenarios.

More relevant are methods modeling dynamic HOI. Many condition on text or action labels~\cite{li2024controllable, diller2024cg, song2024hoianimator,li2024task,cao2025avatargo, xu2026interprior}, which is flexible but lacks fine-grained control and detailed interactions.
Such flexible conditioning is explored for human motion generation~\cite{li2025unimotion, li2026frankenmotion, nazarenus2026actionplan}, but without modeling HOI.
Others, like TRUMANS~\cite{jiang2024scaling}, condition on the scene, improving realism but requiring prior scene knowledge.
Visual conditioning (RGB~\cite{xie2022chore, xie2023vistracker, xie2024InterTrack, fan2023arctic, nam2024joint, zhang2024hoi, xie2024template, tripathi2023deco, dai2024interfusion, antic2026lexis}, multi-view~\cite{jiang2022neuralhofusion,zhang2023neuraldome}, RGBD~\cite{bhatnagar2022behave, huang2024intercap}, or text and images~\cite{yang2024f}) boosts fidelity but relies on cameras, limiting scalability.
I'M HOI~\cite{zhao2024imhoi} uses an object-mounted IMU, enhancing dynamics but requiring the instrumented object.
Several methods use an AMASS motion prior externally to the HOI model via retrieval-and-optimization (InterDreamer~\cite{xu2024interdreamer}) or fine-tuning an AMASS-pretrained MDM (HOI-Diff~\cite{peng2023hoi}), rather than training jointly on motion-only and HOI data.

Another line of research generates interactions from partial information, such as past observations~\cite{ghosh2023imos,xu2023interdiff}, object positions~\cite{kulkarni2024nifty, braun2024physically, li2023object}, or human pose~\cite{petrov2023popup, zhang2024force}.
Among these, TriDi~\cite{petrov2024tridi} models the joint human-object-interaction distribution, but operates on static poses and trains on HOI-only data.
{\methodname} extends the tri-variate formulation to arbitrary-length sequences and, crucially, trains from both motion-only (AMASS) and HOI data in a single model, building the robust motion prior that sparse egocentric conditioning requires.

\subsection{Egocentric HOI Reconstruction Methods}
Most methods that model human-scene interaction from egocentric data assume static environments~\cite{HPS, dai2022hsc4d,lee2024mocap,liu2024egohdm}, and are therefore less relevant to our focus on dynamic interaction.

iReplica~\cite{guzov24ireplica} is the only other method considering both humans and dynamic objects in egocentric scenarios.
However, iReplica has significant limitations: it requires full-body IMUs, a pre-defined 3D scene, and a complex initialization process. Furthermore, it relies on hand-crafted heuristics for interaction modeling, whereas {\methodname} learns HOI dynamics from data.
Recent methods like EgoGrasp~\cite{fu2026egograsp} and WHOLE~\cite{ye2026whole} recover interactions from egocentric video but focus solely on hands; Glove2Hand~\cite{zhang2026glove2hand} synthesizes hand-object interaction from multi-modal sensing gloves, likewise modeling only the hands and requiring specialized worn hardware.
Concurrent work IMU-HOI~\cite{lin2026imu} reconstructs HOI from six body-worn IMUs plus an object-mounted IMU. Unlike {\methodname}, it requires a dense IMU setup and an IMU attached to the object, whereas {\methodname} operates from sparse three-point head-and-hand tracking and uninstrumented objects.
{\methodname} is the first method to model full-body HOI under sparse egocentric conditioning and over arbitrary-length sequences, handling objects of various types and sizes from wearable 3-point trackers, and remaining robust to intermittent hand tracking.

\section{Method} 
\label{sec:method}
In this section, we present {\methodname}, the first approach for joint human-object interaction modeling from head and wrist tracking. 
At the core of our method is a transformer-based diffusion model that predicts human motion $\h$, object motion $\obj$, and contact sequence $\inter$ from three-point conditioning.
Subsequent sections introduce the representations for all modalities (Sec.~\ref{sec:method_repr}) and describe our formulation of the diffusion process, architecture, and training (Sec.~\ref{sec:method_model}). 
Finally, we present the smooth inpainting that we adopt for online inference with arbitrary sequence lengths (Sec.~\ref{sec:method_inference}).

\begin{wrapfigure}{R}{0.298\linewidth}
    \vspace{-1.14cm}
    \footnotesize
    \begin{center}
        \includegraphics[trim=0cm 0cm 0cm 0cm,clip,width=0.96\linewidth]{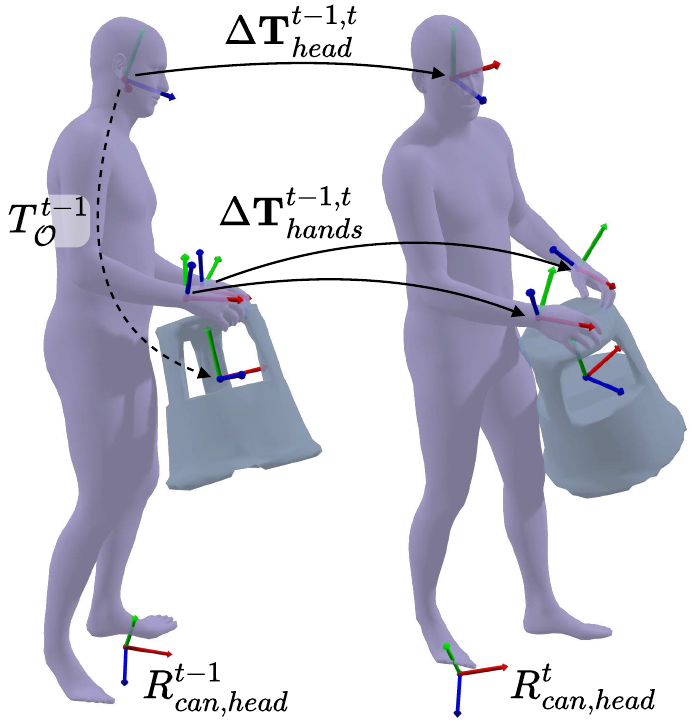}
        \vspace{-4pt}
        \caption{\textbf{Representation}. 
            {\methodname} operates in a per-frame head-centric coordinate system.
        }
        \label{fig:head_centric}
    \end{center}
    \vspace{-1.5cm}
\end{wrapfigure}

\subsection{Representation for Human, Object and Contact}
\label{sec:method_repr}
\subsubsection{Head-centric modeling.}
One of the key design choices for our method is the representation for the network's input and output modalities.
Unlike sequence-level canonicalization~\cite{li2023ego, guzov-jiang2025hmd2}, we build on the per-frame variant of EgoAllo~\cite{yi2025egoallo} by extending the representation to hands and objects. 
We choose to use the canonicalized head position on \textit{each frame} as an anchor for the object's position.
This per-frame definition maintains invariance to the global configuration of the sequence and allows the use of arbitrary chunks of the sequence for inference without explicitly canonicalizing each window.
We illustrate this representation in Fig.~\ref{fig:head_centric} and discuss details in the following paragraphs.

\subsubsection{Object.}
For every object, we assume that its canonical mesh is given to the model as input.
Hence, we represent the object as a sequence of its $SE(3)$ transformations in a head-centric coordinate frame, consisting of rotation and translation pairs w.r.t. the head at each frame $\objglobal=(R_{\obj},t_{\obj})$. 
Following \cite{zhou2019continuity}, we convert all rotations to $\real^6$.

Thus a sequence of $\sequence$ object poses is denoted as:
\begin{equation}
    \obj=\{ \objglobal^{1..\sequence} \}
\end{equation}

To represent the object's class and geometry, we encode the class label in a one-hot encoded vector $\objonehot$, and extract a feature vector $\objfeatures \in \real^{1024}$ from the canonicalized object vertices $\objtemplatev$ using PointNext~\cite{qian2022pointnext}.
The resulting pair $\condobj = (\objonehot, \objfeatures)$ is used as a global conditioning for {\methodname}. 

\subsubsection{Human.}
To represent the human body, SMPL-X~\cite{SMPL-X:2019} is a natural choice. 
SMPL-X is a parametric body model that can be seen as a function $SMPL(\hglobal, \allowbreak \hposefull, \hshape, \hface)$ of root position and orientation $\hglobal \in SE(3)$, pose $\hposefull$, shape $\hshape$, and facial parameters $\hface$.
The function $SMPL$ maps parameters to posed vertices $\htemplatev \in \real^{10475 \times 3}$ of the predefined template mesh. 
The pose vector $\hposefull$ is a concatenation of parameters for body, hands, eyes, and jaw poses in axis-angle format. 
As our method focuses on realistic full-body interactions, we model only the body pose $\hpose \in \real^{21 \times 6}$ part of the pose vector and assume known shape parameters $\hshape$; for the rest of the manuscript, we simplify the notation to $SMPL(\hglobal, \hpose)$.
We follow standard practice and infer $\hglobal$ via aligning the head joint of the SMPL-X model with the head position from three-point tracking.
 
Hence, we define a sequence of human motion with $\sequence$ frames as:
\begin{equation}
    \h = \{ \hpose^{1..\sequence} \}
\end{equation}

\subsubsection{Contacts.} 
Explicit contact modeling is crucial for robust interaction reconstruction, providing strong training signals, enabling self-supervised guidance, and allowing flexible control via sparse conditioning.
We model contact as a continuous modality $\inter$, encompassing both human-object and human-ground interactions.
For human-object contact, we compute the shortest distance $d(p, \objtemplatev)$ from sampled SMPL-X surface points $\interpoints$ to the object.
To fit the diffusion framework and avoid instability, we map distances to $[0, 1]$ using sigmoid:
\begin{equation}
    \intervec^{\text{HOI}} = \left\{ \sigma \left( \alpha \cdot (\tau_c - d(p, \objtemplatev)) \right) \mid p \in \interpoints \subset \htemplatev \right\}
\end{equation}
where $\sigma$ is the sigmoid function, $\tau_c$ is the distance threshold, and $\alpha$ controls decay sharpness.
Similarly, we compute human-environment contact $\intervec^{\text{Env}}$ for lower body joints based on velocity and ground proximity~\cite{rempe2021humor, yi2025egoallo}.
The final contact vector is $\intervec = \{\intervec^{\text{HOI}}, \intervec^{\text{Env}}\}$, and the sequence is defined as:
\begin{equation}
    \inter=\{ \intervec^{1..\sequence} \}
\end{equation}

\subsubsection{Egocentric conditioning.}
{\methodname} extends the conditioning formulation of EgoAllo \cite{yi2025egoallo} by incorporating relative hand transformations at each frame into the method's conditioning.
The method is conditioned on canonicalized head and both hands orientations $\egorotcan, \egorothandscan$, head-to-floor distance $\egoheight$, and relative head and hand transformations.
The relative head transformation between the current and previous frame $\egoheaddeltat \in SE(3)$ is computed as:
\begin{equation}
    \egoheaddeltat = (\mat{T}^{t-1}_{\text{world, head}})^{-1} \cdot \mat{T}^{t}_{\text{world, head}}
\end{equation}
where $\mat{T}^{t}_{\text{world, head}}$ is the transformation of the head at time $t$.
Similarly, we compute the relative transformation for the hands $\egohanddeltat$. We represent the transformation as $\real^6$ rotation and $\real^3$ translation before passing it to the network.

Thus, egocentric conditioning for a sequence $\condego$ is defined as:
\begin{equation}
    \condego = \{ [\egoheaddeltat, \egorotcan, \egoheight, \egohanddeltat, \egorothandscan]^{1..\sequence} \}
\end{equation}

\subsection{{\methodname} model}
\label{sec:method_model}

{\methodname} models the joint distribution of human motion $\h$, object motion $\obj$, and contact sequence $\inter$, conditioned on the three-point tracking $\condego$ and object features $\condobj$.
Below, we formulate a three-variate diffusion process that is used to model the three modalities within one network and provide details on the underlying network's architecture.

\subsubsection{Background}
A diffusion process in the context of generative neural networks is divided into two phases. 
The forward phase progressively adds noise to an original data sample, while the backward phase uses a learned model to recover the sample from noise.
We adopt the formulation of Denoising Diffusion Probabilistic Model (DDPM)~\cite{ho2020denoising} in our work with a modification following~\cite{ramesh2022hierarchical}, predicting the original sample with the neural network instead of predicting the added noise.
To achieve this, we parametrize the reverse process by a denoising neural network $\mathcal{D}_\psi$ that is trained to recover the original sample $\mathbf{z}^0$ from the noisy sample $\mathbf{z}^\set{T}$ at denoising step $\set{T}$ given the condition $c$. Defining for brevity 
$\mathbb{E}_p \equiv \mathbb{E}_{\mathbf{z}^0 \sim p_{data}}$, 
$\mathbb{E}_\set{T} \equiv \mathbb{E}_{\set{T} \sim \mathcal{U}\{0,...,T\}}$, and 
$\mathbb{E}_q \equiv \mathbb{E}_{\mathbf{z}^\set{T} \sim q(\mathbf{z}^\set{T} | \mathbf{z}^0)}$ 
we obtain the training objective (the full definition of forward and backward processes is included in the Sup. Mat.):
\begin{equation}
    \begin{aligned}
        \min_\psi  
           \mathbb{E}_p\, \mathbb{E}_\set{T}\, \mathbb{E}_q\, \|\mathcal{D}_\psi(\mathbf{z}^\set{T}; c, \set{T}) - \mathbf{z}^0\|_2.
    \end{aligned}
    \label{eqn:objective}
\end{equation}

The original formulation of the diffusion model~\cite{sohl2015deep, ho2020denoising} focuses on generating single-modality data, e.g., images.
Inspired by TriDi~\cite{petrov2024tridi}, we formulate a tri-variate diffusion process for HOI modeling, proposing a new formulation that diffuses motion sequences.

\begin{figure}[t!]
    \centering
    \scriptsize
    \includegraphics[trim=0cm 0cm 0cm 0cm,clip,width=1.0\linewidth]{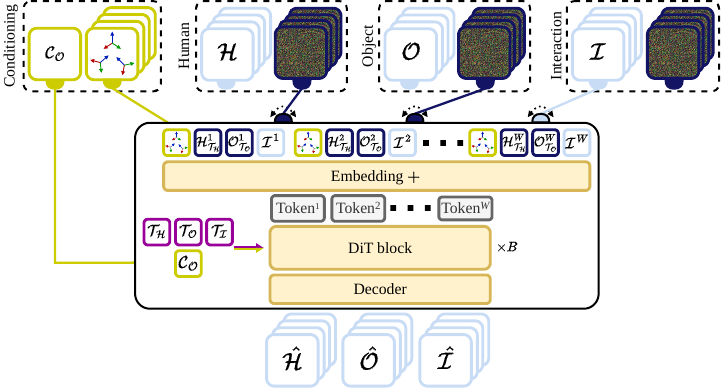}
    \caption{\textbf{{\methodname} overview.} 
    {\methodname} requires just head and hand tracking and object class, to predict Human, Object, and Interaction. 
    The input tokens are composed of {\color{cond_col}{\bf{condition}}}, and of either {\color{gt_col}{\bf{observed modality}}}, or {\color{noise_col}{\bf{noise}}} for $\h, \obj, \text{and } \inter$. 
    For every modality, we use a unique denoising {\color{t_col}{\bf{step}}}. 
    Our model allows flexible input configuration.
    In the example above we use contacts $\inter$ as an {\color{gt_col}{\bf{additional input}}} to the network, that infers the {\color{noise_col}{\bf{other modalities}}} $\h \text{ and } \obj$, matching the extended condition.
    }
    \vspace{-10pt}
    \label{fig:overview}
\end{figure}

\subsubsection{Tri-variate diffusion with independent schedules.}
We formulate a three-variate diffusion process for human motion $\h$, object trajectory $\obj$, and sequence of contacts $\inter$, denoting the corresponding sets of denoising steps as $\thuman, \tobj, \tinter$. We define:
\begin{equation}
    \begin{aligned}
    \mathbb{E}_p &\equiv \mathbb{E}_{(\h^0, \obj^0, \inter^0) \sim p(\h,\obj, \inter | \condego)},\\
    \mathbb{E}_\set{T} &\equiv \mathbb{E}_{(\thuman, \tobj, \tinter) \sim \mathcal{U}\{0,...,T\}^{N\times3}},\\
    \mathbb{E}_q &\equiv \mathbb{E}_{\h^{\thuman} \sim q(\h^{\thuman} | \h^0),
                                \obj^{\tobj} \sim q(\obj^{\tobj} | \obj^0),
                                \inter^{\tinter} \sim q(\inter^{\tinter} | \inter^0)}
    \end{aligned}
\end{equation}

The key feature of this formulation is that {\methodname} diffuses the three modalities following three independent time schedules, allowing them to vary (e.g., one can provide tracking information for the human in addition to three-point conditioning and predict the object motion and contact sequence corresponding to it).
The main minimization objective for parameters $\methodparams$ of a model $\methodnetwork$ is:
\vspace{-15pt}
\begin{equation}
    \begin{aligned}
    \mathbb{E}_p \mathbb{E}_\set{T} \mathbb{E}_q
        \| & \methodnetwork(
            \h^{\thuman}, \obj^{\tobj}, \inter^{\tinter} ; 
            \thuman, \tobj, \tinter; \condobj, \condego)
            - (\h^0, \obj^0, \inter^0) \|_2
    \label{eqn:diffusion_objective}
    \end{aligned}
\end{equation}

\subsubsection{Universal multi-modal architecture.}
We build the {\methodname} denoising network on the Diffusion Transformer (DiT)~\cite{peebles2023scalable} with rotary positional embeddings~\cite{su2024roformer}, adapting it to our multi-modal setting (Fig.~\ref{fig:overview}).
The model takes as input the sequences $\h$, $\obj$, and $\inter$ with varying noise levels, along with egocentric $\condego$ and object $\condobj$ conditioning, and the denoising step for each modality $\{ \thuman, \tobj, \tinter \}$.
A key advantage of our design is the use of independent noise schedules for each modality, unlike standard approaches that use a single schedule.
This enables flexible conditioning: by setting the noise level to zero for known modalities (e.g., observed human motion), the model effectively predicts the remaining components (e.g., object motion and contacts) consistent with the input.
When no interaction data is available, it generates realistic motion based solely on egocentric conditioning.
Furthermore, this formulation naturally handles partial observations, allowing {\methodname} to leverage sparse signals -- such as intermittent object tracking or partial human pose from IMUs -- to constrain the generation.
Ultimately, the method outputs the full human-object interaction sequence $\{ \hat{\h},\hat{\obj},\hat{\inter} \}$.

\begin{figure}[t!]
    \centering
    \includegraphics[trim=0cm 0cm 0cm 0cm,clip,width=1.0\linewidth]{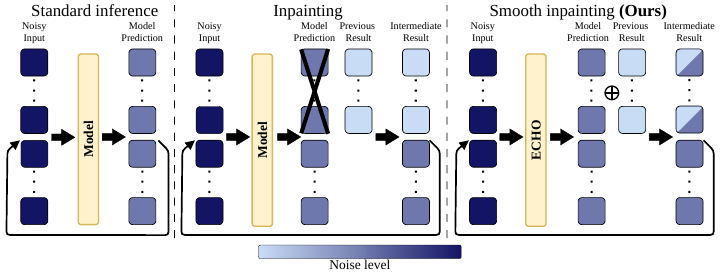}
    \caption{\textbf{Comparison of inference strategies.} Standard per-window inference (left) ignores the context of the past predictions. Inpainting (middle) uses past prediction as condition but drops new predictions for the overlapping region. Our smooth inpainting (right) blends past and current predictions in the overlapping region on every diffusion step, ensuring seamless transitions.}
    \vspace{-10pt}
    \label{fig:smooth_inpainting}
\end{figure}

\subsubsection{Training.}
During training, each modality ($\h$, $\obj$, $\inter$) is either diffused or provided as clean condition. 
To ensure stability, we sample from the $2^3$ combinations of noisy/denoised states and synchronize the noise level across all diffused modalities, instead of sampling independent noise levels for each modality. 
This effectively trains the model to reconstruct missing modalities from the observed ones.
To improve generalization, we simulate intermittent tracking by randomly dropping hand and object conditioning. 
Experiments confirm that {\methodname} remains robust to moderate tracking degradation.

The objective function is a weighted sum of six terms: reconstruction losses for each diffused modality, object trajectory smoothness, human joint error, and a foot skating penalty (details in Sup. Mat.).
Given the limited scale of HOI datasets (OMOMO, BEHAVE), we augment training with the large-scale AMASS~\cite{AMASS:ICCV:2019} dataset to learn a robust human motion prior. 
For AMASS samples, we replace object conditioning $\condobj$ with learnable tokens, signaling the model to ignore object interactions; ablations confirm the benefits of this joint training.

\begin{figure*}[t!]
    \centering
    \includegraphics[trim=0cm 0cm 0cm 0cm,clip,width=1.0\linewidth]{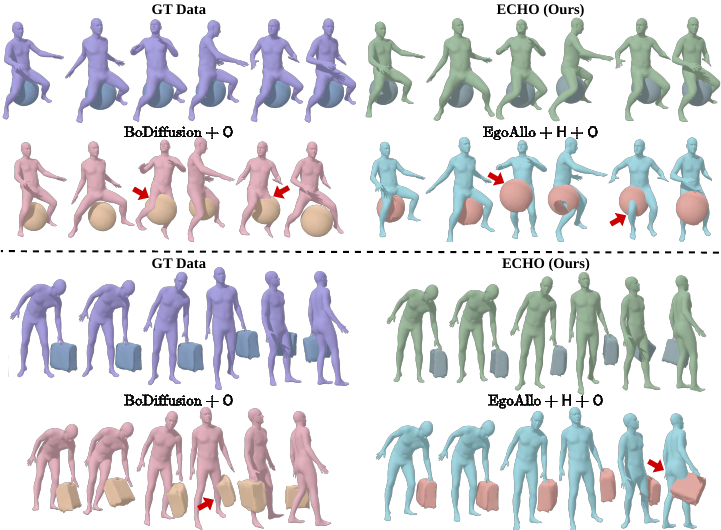}
    \caption{\textbf{Qualitative results of {\methodname}}. Our method accurately reconstructs human-object interactions across diverse scenarios. In contrast, competing methods often fail to capture correct contact dynamics, leading to artifacts such as object penetration or floating. For dynamic visualizations, please refer to the supplementary video.}
    \vspace{-10pt}
    \label{fig:results}
\end{figure*}

\subsection{Inference}
\label{sec:method_inference}
\subsubsection{Smooth inpainting}
To generate temporal sequences, recent methods~\cite{yi2025egoallo} employ a sliding window approach. 
Unlike offline methods~\cite{tevet2022human, castillo2023bodiffusion}, this approach doesn't require the full sequence to be present at the start, enabling online inference.
However, sliding windows require post-processing to stitch results.
HMD\textsuperscript{2}~\cite{guzov-jiang2025hmd2} addresses this via inpainting-based inference, conditioning new windows on past predictions. 
However, it discards new predictions in the overlapping region, relying solely on past data. 
Our idea is to take the inpainting-based inference one step further by incorporating smooth blending between past and new predictions, which we call \textit{smooth inpainting} (Fig.~\ref{fig:smooth_inpainting}).
It replaces predictions in the overlapping region with a weighted average of past and current predictions, allowing for a gradual transition between windows.
This enables {\methodname} to autoregressively generate arbitrarily long sequences with smooth transitions. 
Additionally, the overlap size can be tuned to balance context length and inference latency for real-time applications.

\subsubsection{Inference guidance}
To further enhance the quality of generated interactions, we adopt a classifier-based guidance~\cite{dhariwal2021diffusion} approach at inference time. 
Based on the guidance function's value, the network's prediction is updated at each step of the denoising process.
We formulate the guidance loss to ensure that the predicted human and object meshes align with the predicted contact vector $\predintervec$. 
The function, therefore, includes two terms: one for human-object contact and one for foot-floor contact.

\section{Experiments}
\label{sec:experiments}
\begin{figure*}[t!]
    \centering
    \includegraphics[width=1.0\linewidth]{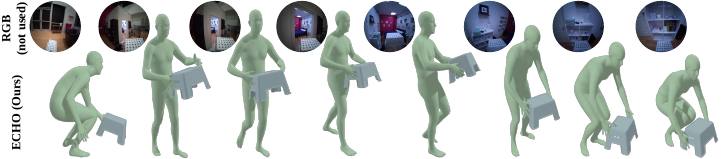}
    \caption{\textbf{Qualitative results of {\methodname}}. We demonstrate generalization to novel motion and objects from the Aria Digital Twin~\cite{pan2023aria}; RGB is included for reference.}
    \vspace{-10pt}
    \label{fig:adt}
\end{figure*}

\subsubsection{Implementation details.} The {\methodname} model comprises 57.7M parameters and is trained using the AdamW optimizer~\cite{loshchilov2017decoupled} for 300k steps (approximately 30 hours) with a batch size of 256 and a learning rate of $5e^{-4}$ on a single NVIDIA RTX 5090 GPU.
For inference, we use 100 DDPM steps to balance generation quality and latency.
With a 30-frame overlap, {\methodname} processes a 60-frame window in 640 ms without guidance and 980 ms with guidance on an RTX 5090 GPU. 
This corresponds to a throughput of approximately 46 FPS and 30 FPS, respectively, enabling operation at low latency of 640--980\,ms per window, \ie, $\sim$21--33\,ms amortized per frame.
Visualizations are generated using aitviewer~\cite{Kaufmann_Vechev_aitviewer_2022} and Blendify~\cite{blendify2024}.

\subsubsection{Datasets.}
We train and evaluate our method on the union of the BEHAVE~\cite{bhatnagar2022behave}, OMOMO~\cite{li2023object}, and AMASS~\cite{AMASS:ICCV:2019} datasets. We follow the official train-test splits for BEHAVE and OMOMO, and the EgoAllo~\cite{yi2025egoallo} split for AMASS. To use the SMPL-X body model in {\methodname}, we convert BEHAVE sequences from the SMPL+H~\cite{MANO:SIGGRAPHASIA:2017} format using code from~\cite{sarandi24nlf}.
We downsample all sequences to 30 fps for consistency, as the BEHAVE data is limited to this frame rate. During training, we sample windows of size $W=60$. For evaluation, we perform continuous inference on full sequences.

\subsubsection{Metrics.}
To evaluate human motion reconstruction, we use the Mean Per-Joint Position Error (MPJPE, in cm), Mean Per-Joint Velocity Error (MPJVE \cite{zheng2023realistic}), and Foot Contact (FC) score~\cite{li2023ego, yi2025egoallo}.
For object reconstruction, we compute the vertex-to-vertex error $E_{v2v}$~\cite{petrov2023popup, xu2023interdiff} (cm), center error $E_c$~\cite{petrov2023popup} (cm), and contact accuracy $\text{Acc}_{\inter}$~\cite{petrov2024tridi}.
Detailed metric definitions are provided in the Supp. Mat.

\subsection{Comparison with baselines}

\renewcommand{\arraystretch}{1.2}
\begin{table}[tb]
    \setlength{\tabcolsep}{2pt}
    \centering
    \scriptsize
    \caption{\textbf{Comparison with baselines on BEHAVE and OMOMO}. {\methodname} demonstrates better performance for human-object interaction modeling and competitive motion modeling quality.}
    \vspace{-5pt}
    \label{tab:baselines}
    \begin{tabular}{l@{\hspace{0.5em}}ccccccccc}
    \specialrule{.1em}{.05em}{.05em} 
        \multicolumn{9}{c}{\textbf{BEHAVE}} \\
        \multirow{3}{*}{\bf Method} & 
            \multicolumn{3}{c}{\bf Human} & & 
            \multicolumn{4}{c}{\bf Object}  \\
        \cmidrule{2-4} \cmidrule{5-9}
            & MPJPE$\downarrow$ & MPJVE$\downarrow$ & FC$\uparrow$ & &
            $E_{v2v}\downarrow$ & $E_{c}\downarrow$ & Rot. Diff.$\downarrow$ & $\text{Acc}_{\inter}\uparrow$ \\       
        \cmidrule{2-4} \cmidrule{5-9}
        Data                                                &
            - & - & 0.73 & &
            - & - & - & - \\
        \hline
        \textbf{\bodiffo[\cite{castillo2023bodiffusion}]} &
            ${8.3^{\pm0.2}}$ & ${10.1^{\pm0.3}}$ & ${0.82^{\pm0.04}}$ & &
            ${44.2^{\pm1.2}}$ & ${29.9^{\pm1.2}}$ & $\second{110.9^{\pm4.4}}$ & $\second{91.8^{\pm0.4}}$ \\
        \textbf{\egoalloho[\cite{yi2025egoallo}]}    &
            $\second{7.6^{\pm0.1}}$ & $\second{8.6^{\pm0.2}}$ & $\first{0.95^{\pm0.01}}$ & &
            $\second{39.1^{\pm1.1}}$ & $\second{22.5^{\pm0.7}}$ & ${111.6^{\pm3.8}}$ & ${91.7^{\pm0.3}}$ \\
        \hline
        {\methodname} \textbf{(Ours)}                       & 
            $\first{6.8^{\pm0.1}}$ & $\first{7.5^{\pm0.1}}$ & $\second{0.94^{\pm0.00}}$ & &
            $\first{33.5^{\pm0.5}}$ & $\first{20.1^{\pm0.3}}$ & $\first{92.9^{\pm2.2}}$ & $\first{93.1^{\pm0.2}}$ \\
    \specialrule{.1em}{.05em}{.05em} 
    \end{tabular}

    \begin{tabular}{l@{\hspace{0.5em}}cccccccc}
    \specialrule{.1em}{.05em}{.05em} 
        \multicolumn{9}{c}{\textbf{OMOMO}} \\
        \multirow{3}{*}{\bf Method} & 
            \multicolumn{3}{c}{\bf Human} & & 
            \multicolumn{4}{c}{\bf Object}  \\
        \cmidrule{2-4} \cmidrule{5-9}
            & MPJPE$\downarrow$ & MPJVE$\downarrow$ & FC$\uparrow$ & &
            $E_{v2v}\downarrow$ & $E_{c}\downarrow$ & Rot. Diff.$\downarrow$ & $\text{Acc}_{\inter}\uparrow$ \\       
        \cmidrule{2-4} \cmidrule{5-9}
        Data                                                &
            - & - & 0.96 & &
            - & - & - & - \\
        \hline
        \textbf{\bodiffo[\cite{castillo2023bodiffusion}]}    &
            ${7.6^{\pm0.4}}$ & ${8.6^{\pm0.3}}$ & $\first{0.98^{\pm0.01}}$ & &
            ${33.2^{\pm1.9}}$ & ${22.2^{\pm1.7}}$ & $\second{94.9^{\pm7.4}}$ & $\second{96.2^{\pm0.3}}$ \\
        \textbf{\egoalloho[\cite{yi2025egoallo}]}    &
            $\second{6.6^{\pm0.1}}$ & $\second{7.3^{\pm0.1}}$ & $\second{0.95^{\pm0.01}}$ & &
            $\second{30.8^{\pm0.9}}$ & $\second{18.3^{\pm0.5}}$ & ${98.2^{\pm5.3}}$ & ${96.5^{\pm0.2}}$ \\
        \hline
        {\methodname} \textbf{(Ours)}                       & 
            $\first{6.0^{\pm0.1}}$ & $\first{6.1^{\pm0.1}}$ & ${0.93^{\pm0.01}}$ & &
            $\first{26.5^{\pm1.1}}$ & $\first{15.2^{\pm0.3}}$ & $\first{86.5^{\pm6.7}}$ & $\first{96.9^{\pm0.1}}$ \\
    \specialrule{.1em}{.05em}{.05em} 
    \end{tabular}
    \vspace{-5pt}
\end{table}
\renewcommand{\arraystretch}{1.0}

\subsubsection{Baselines.}
{\methodname} is the first approach for the end-to-end modeling of egocentric human-object interactions.
To evaluate its performance, we select several existing end-to-end egocentric motion modeling methods as baselines and extend them for HOI modeling.
While some recent methods employ multi-stage training and inference, adapting them to HOI modeling would require significant architectural modifications.
We compare against \textbf{BoDiffusion}~\cite{castillo2023bodiffusion}, which builds on DiT~\cite{peebles2023scalable} to process concatenated input tracking and noisy motion.
During training, we canonicalize each window to a head-centric space to ensure a fair comparison.
To evaluate HOI prediction, we extend BoDiffusion to include object modeling, referring to this baseline as \textbf{\bodiffo}.
We modify the network to process object poses and conditioning by concatenating object transformations to the human motion and egocentric tokens, and appending object shape embeddings and class information to the global conditioning.
We also construct an HOI baseline on top of \textbf{EgoAllo}~\cite{yi2025egoallo} by integrating object pose prediction.
Since EgoAllo is originally conditioned only on head tracking, we extend it to support hand conditioning, following the official implementation.
We refer to this method as \textbf{\egoalloho}.
For fair comparison, we train {\methodname} and all baselines on the union of AMASS, BEHAVE, and OMOMO. 
Additionally, both baselines predict object pose in a head-centric space, matching {\methodname}, to ensure a consistent coordinate system.

\renewcommand{\arraystretch}{1.2}
\begin{table}[t!]
    \setlength{\tabcolsep}{3pt}
    \centering
    \scriptsize

    \caption{\textbf{Quality of motion generation}. {\methodname} outperforms the baselines on the AMASS dataset, demonstrating that our joint HOI formulation effectively learns a strong human motion prior. The significant drop in performance for \textbf{NoAMASS} confirms the importance of training on large-scale motion data.}
    \label{tab:amass}
    
    \begin{tabular}{lccc}
    \specialrule{.1em}{.05em}{.05em}
        \multicolumn{4}{c}{\textbf{AMASS}} \\
        \cmidrule{2-4}
        \textbf{Method} & MPJPE$\downarrow$ & MPJVE$\downarrow$ & FC$\uparrow^{1.0}$ \\
        \cmidrule{2-4}
        \textbf{\bodiffo[\cite{castillo2023bodiffusion}]}       &
            $11.4^{\pm0.3}$ & $14.3^{\pm0.3}$ & $1.0$ \\
        \textbf{\egoalloho[\cite{yi2025egoallo}]}    &
            $\second{8.9^{\pm0.1}}$ & $\second{11.6^{\pm0.1}}$ & $1.0$ \\
        \hline
        {\methodname} \textbf{(Ours)} - NoAMASS             & 
            $43.1^{\pm0.1}$ & $40.3^{\pm0.1}$ & $0.8$ \\ 
        \hline
        {\methodname} \textbf{(Ours)}             & 
            $\first{7.4^{\pm0.1}}$ & $\first{8.6^{\pm0.2}}$ & $1.0$ \\ 
    \specialrule{.1em}{.05em}{.05em} 
    \end{tabular}
    \vspace{-5pt}
\end{table}
\renewcommand{\arraystretch}{1.0}

\subsubsection{HOI generation.}
We evaluate HOI reconstruction quality on the BEHAVE and OMOMO test sets, reporting mean and variance across three runs in Tab.~\ref{tab:baselines}.
{\methodname} significantly outperforms {\bodiffo} in both human and object predictions, and performs on par with {\egoalloho} in human motion while surpassing it in object prediction.
This superior object tracking demonstrates the efficacy of our tri-variate modeling.
Qualitative results (Fig.~\ref{fig:results}) confirm this advantage: while the baselines often fail to maintain realistic contact, resulting in artifacts like penetration or floating, {\methodname} generates physically plausible interactions.
We further demonstrate generalization on an Aria Digital Twin~\cite{pan2023aria} sequence in Fig.~\ref{fig:adt}; see the supplementary video for dynamic visualizations.

\subsubsection{Motion generation.}
While {\methodname} is designed for human-object interactions, robust human motion reconstruction is also essential.
We evaluate motion quality by comparing {\methodname} against the baselines on the AMASS dataset.
Results are reported in Tab.~\ref{tab:amass} (mean and variance across three runs).
Our method outperforms both baselines, demonstrating the effectiveness of our joint modeling formulation.
Notably, performance significantly degrades when {\methodname} is trained without the AMASS dataset (\textbf{NoAMASS}), highlighting the importance of large-scale motion data for learning a strong human motion prior.
For fair comparison, all models (except (\textbf{NoAMASS})) were trained on the union of the three datasets.

\renewcommand{\arraystretch}{1.2}
\begin{table}[t!]
    \setlength{\tabcolsep}{4pt}
    \centering
    \scriptsize
    \caption{\textbf{Evaluation of {\methodname} with noise simulation}. We demonstrate the robustness of {\methodname} to intermittent hand tracking by randomly dropping a percentage of the input. The model maintains stable performance even with significant missing hand tracking data, confirming its resilience to sensor noise.}
    \label{tab:noise}
    
    \begin{tabular}{lccccc}
    \specialrule{.1em}{.05em}{.05em} 
        \multicolumn{6}{c}{\textbf{BEHAVE}} \\
            \multirow{2}{*}{\bf \%} & 
            \multicolumn{2}{c}{\bf Human} & & 
            \multicolumn{2}{c}{\bf Object}  \\
        \cmidrule{2-3} \cmidrule{5-6}
            & MPJPE$\downarrow$ & MPJVE$\downarrow$ & &
            $E_{v2v}\downarrow$ & $E_{c}\downarrow$ \\       
        \cmidrule{2-3} \cmidrule{5-6}
        \textit{0} & 
            ${6.8^{\pm0.1}}$ & ${7.5^{\pm0.1}}$ & &
            ${33.5^{\pm0.5}}$ & ${20.1^{\pm0.3}}$ \\
        \hline
        \textit{25} & 
            ${6.9^{\pm0.1}}$ & ${7.5^{\pm0.2}}$ & &
            ${33.8^{\pm0.8}}$ & ${20.6^{\pm0.5}}$ \\
        \textit{50} &
            ${7.0^{\pm0.2}}$ & ${7.8^{\pm0.2}}$ & &
            ${33.9^{\pm1.0}}$ & ${20.8^{\pm0.6}}$ \\
        \textit{75} &
            ${7.6^{\pm0.3}}$ & ${10.3^{\pm0.5}}$ & &
            ${34.6^{\pm1.3}}$ & ${21.2^{\pm1.0}}$ \\
        \textit{90} &
            ${9.3^{\pm0.5}}$ & ${10.3^{\pm0.6}}$ & &
            ${36.8^{\pm2.0}}$ & ${24.6^{\pm1.9}}$ \\
    \specialrule{.1em}{.05em}{.05em} 
    \end{tabular}
\end{table}
\renewcommand{\arraystretch}{1.0}

\subsection{Noisy conditioning and sparse input}
Egocentric human-object interactions are captured using diverse technologies and settings, often involving data streams (e.g., IMUs, head-mounted cameras) that provide additional, albeit often sparse and noisy, information.
To study the robustness and versatility of {\methodname} under such conditions, we simulate two scenarios: noisy hand tracking and access to additional sparse tracking information.
To simulate intermittent hand tracking, we randomly drop a percentage of hand tracking data from the input while keeping head tracking intact.
Tab.~\ref{tab:noise} demonstrates that {\methodname} is robust to missing hand tracking, showing significant performance degradation only when 75\% or more of the data is missing.

While tracking objects from an egocentric camera remains challenging~\cite{banerjee2025hot3d, zhao2024instance}, the object's location and orientation can often be determined for a few frames.
For human motion, it is also possible to obtain partial tracking from RGB-based pose estimation or IMU suits.
Such information provides important cues to reduce uncertainty, and {\methodname} is designed to leverage these opportunities.
We test this capability by simulating a scenario where either the human or object modality is only partially observed.
Tab.~\ref{tab:sparse} shows that additional constraints significantly improve the accuracy of the partially observed modality, while predictions for the unobserved modality improve slightly.

\renewcommand{\arraystretch}{1.2}
\begin{table}[t!]
    \setlength{\tabcolsep}{4pt}
    \centering
    \scriptsize
    \caption{\textbf{Evaluation of {\methodname} with sparse tracking}. We demonstrate the versatility of {\methodname} by providing additional sparse tracking information alongside egocentric conditioning. Providing partial information for one modality (Human or Object) significantly improves its reconstruction quality and helps regularize the other.}
    \label{tab:sparse}
    
    \begin{tabular}{lcccccc}
    \specialrule{.1em}{.05em}{.05em} 
        \multicolumn{7}{c}{\textbf{BEHAVE}} \\
        \multirow{2}{*}{\bf Mode} & 
            \multirow{2}{*}{\bf \%} & 
            \multicolumn{2}{c}{\bf Human} & & 
            \multicolumn{2}{c}{\bf Object}  \\
        \cmidrule{3-4} \cmidrule{6-7}
            & & MPJPE$\downarrow$ & MPJVE$\downarrow$ & &
            $E_{v2v}\downarrow$ & $E_{c}\downarrow$ \\       
        \cmidrule{3-4} \cmidrule{6-7}
        \multicolumn{2}{l}{\bf {\methodname}} & 
            ${6.82^{\pm0.08}}$ & ${7.50^{\pm0.11}}$ & &
            ${33.46^{\pm0.50}}$ & ${20.13^{\pm0.26}}$ \\
        \hline
        \multirow{4}{*}{\rottext[90]{\makecell{\hphantom{AAAA}Sparse $\h$\\\hphantom{AAAA}available}}} &
         \textit{10} &
            ${5.84^{\pm0.10}}$ & ${6.27^{\pm0.12}}$ & &
            ${33.44^{\pm0.73}}$ & ${20.12^{\pm0.38}}$ \\
        & \textit{25} &
            ${4.84^{\pm0.09}}$ & ${5.09^{\pm0.12}}$ & &
            ${33.42^{\pm0.74}}$ & ${20.00^{\pm0.38}}$ \\
        & \textit{50} &
            ${3.71^{\pm0.07}}$ & ${3.80^{\pm0.10}}$ & &
            ${33.29^{\pm0.72}}$ & ${19.81^{\pm0.37}}$ \\
        & \textit{100} &
            - & - & &
            ${32.79^{\pm0.65}}$ & ${19.41^{\pm0.31}}$ \\
        \hline
        \multirow{4}{*}{\rottext[90]{\makecell{\hphantom{AAAA}Sparse $\obj$\\\hphantom{AAAA}available}}} &
         \textit{10} &
            ${6.81^{\pm0.10}}$ & ${7.48^{\pm0.12}}$ & &
            ${27.20^{\pm0.70}}$ & ${16.55^{\pm0.39}}$ \\
        & \textit{25} &
            ${6.81^{\pm0.09}}$ & ${7.47^{\pm0.11}}$ & &
            ${19.77^{\pm0.78}}$ & ${12.69^{\pm0.42}}$ \\
        & \textit{50} &
            ${6.79^{\pm0.09}}$ & ${7.45^{\pm0.11}}$ & &
            ${10.75^{\pm0.64}}$ & ${7.93^{\pm0.38}}$ \\
        & \textit{100} &
            ${6.69^{\pm0.08}}$ & ${7.37^{\pm0.10}}$ & & 
            - & - \\
    \specialrule{.1em}{.05em}{.05em} 
    \end{tabular}
\end{table}
\renewcommand{\arraystretch}{1.0}

\subsection{Ablation}
We conduct an ablation study to analyze the effectiveness of our method's components.
All models are trained in the same setting, varying only the ablated component.
First, we evaluate the model without inference-time guidance (\textbf{NoGuide}), observing that it primarily affects object prediction quality on noisier data (i.e., BEHAVE).
Results on BEHAVE are presented in Tab.~\ref{tab:abl_behave}; OMOMO results are in the Supplementary Material.
\textbf{Inpaint w/o smooth} performs inference using standard inpainting, without smooth blending between past and current predictions.
The performance drop without smooth inpainting highlights its importance for generating temporally consistent interactions.
The model without the interaction modality (i.e., using only $\boldsymbol{(\h, \obj)}$) exhibits substantially worse quality for both human and object motion, suggesting that contacts play a crucial role in linking these modalities.
Training without the AMASS dataset (\textbf{NoAMASS}) leads to a significant decrease in human motion modeling quality, once again highlighting the importance of large-scale datasets for learning human motion priors.

\renewcommand{\arraystretch}{1.2}
\begin{table}[t!]
    \setlength{\tabcolsep}{3pt}
    \centering
    \scriptsize

    \caption{\textbf{Ablation study on BEHAVE.} Evaluating the impact of {\methodname} components proves the usefulness of guidance, smooth inpainting, usage of three modalities, and training with AMASS data.}
    \label{tab:abl_behave}
    
    \begin{tabular}{lccccc}
    \specialrule{.1em}{.05em}{.05em} 
        \multicolumn{6}{c}{\textbf{BEHAVE}} \\
        \multirow{2}{*}{\bf Method} & 
            \multicolumn{2}{c}{\bf Human} & & 
            \multicolumn{2}{c}{\bf Object}  \\
        \cmidrule{2-3} \cmidrule{5-6}
            & MPJPE$\downarrow$ & MPJVE$\downarrow$ & &
            $E_{v2v}\downarrow$ & $E_{c}\downarrow$ \\
        \cmidrule{2-3} \cmidrule{5-6}
        {\methodname}                                 & 
            ${6.8^{\pm0.1}}$ & ${7.5^{\pm0.1}}$ & &
            ${33.5^{\pm0.5}}$ & ${20.1^{\pm0.3}}$ \\
        \hline
        NoGuide                 &
            ${6.8^{\pm0.1}}$ & ${7.5^{\pm0.1}}$ & &
            ${33.6^{\pm0.5}}$ & ${20.3^{\pm0.4}}$ \\
        Inpaint w/o smooth      &
            ${6.9^{\pm0.1}}$ & ${7.6^{\pm0.1}}$ & &
            ${33.7^{\pm0.5}}$ & ${20.4^{\pm0.3}}$ \\
        ($\h$, $\obj$)          &
            ${8.1^{\pm0.1}}$ & ${9.0^{\pm0.1}}$ & &
            ${34.4^{\pm0.3}}$ & ${20.7^{\pm0.1}}$ \\
        NoAMASS                 &
            ${8.7^{\pm0.1}}$ & ${9.6^{\pm0.1}}$ & &
            ${34.7^{\pm0.2}}$ & ${21.8^{\pm0.2}}$ \\
    \specialrule{.1em}{.05em}{.05em} 
    \end{tabular}
    \vspace{-10pt}
\end{table}
\renewcommand{\arraystretch}{1.0}

\section{Conclusions}
\label{sec:conclusions}
In this work, we introduced {\methodname}, the first unified framework for modeling human-object interactions solely from sparse egocentric tracking. 
Central to our approach is a novel tri-variate diffusion formulation with independent noise schedules that jointly models human pose, object motion, and contact dynamics. 
This design not only captures the complex interdependencies between the user and the object but also enables flexible inference, allowing the model to adapt to intermittent tracking and partial observations. 
Crucially, it facilitates training on a combination of large-scale human motion datasets and smaller HOI collections, learning strong priors while capturing interaction nuances.
Additionally, our smooth inpainting inference mechanism ensures temporally consistent generation for sequences of arbitrary length. 
Extensive evaluations demonstrate that {\methodname} significantly outperforms state-of-the-art methods, offering a robust and versatile solution for egocentric HOI reconstruction.

\subsubsection{Limitations and future work.}
Despite its strong performance, {\methodname} has limitations that open exciting directions for future research.
First, our current model focuses on object interactions and environmental contacts with the ground.
Incorporating sophisticated constraints (\eg, those coming from dynamic surroundings) would be crucial for consistent long-term motion in complex scenes.
Second, although robust for various object types, the absence of fine-grained finger tracking limits the reconstruction of dexterous interactions with small objects (e.g., pens, scissors). 
Integrating additional modalities, such as egocentric RGB video, could help resolve these fine motor details. 
Finally, extending our framework to support visual conditioning, alongside the collection of currently absent RGB-based egocentric HOI datasets, represents a promising avenue for achieving comprehensive egocentric perception.

\subsubsection{Acknowledgments}
{
Special thanks to Nikita Kister and Berna Kabadayi for the helpful discussions. 
This work is funded by the Deutsche Forschungsgemeinschaft - 409792180 (EmmyNoether Programme, project: Real Virtual Humans). 
G. Pons-Moll is a member of the Machine Learning Cluster of Excellence, EXC number 2064/1 – Project number 390727645. 
The authors thank the International Max Planck Research School for Intelligent Systems (IMPRS-IS) for supporting I.~A.~Petrov. 
R. Marin has been supported by the European Union’s Horizon 2020 research and innovation program under the Marie Skłodowska-Curie grant agreement No 101109330. 
The project was made possible by funding from the Carl Zeiss Foundation.
The computational resources for this project were provided by the Google Cloud grant.
This work was supported by the European Research Council (ERC) Advanced Grant SIMULACRON and by the GNI Project “AI4Twinning”. 
}

\clearpage
{
    \small
    \bibliographystyle{splncs04}
    \bibliography{main}

\begin{thebibliography}{100}
\providecommand{\url}[1]{\texttt{#1}}
\providecommand{\urlprefix}{URL }
\providecommand{\doi}[1]{https://doi.org/#1}

\bibitem{akada2022unrealego}
Akada, H., Wang, J., Shimada, S., Takahashi, M., Theobalt, C., Golyanik, V.: Unrealego: A new dataset for robust egocentric 3d human motion capture. In: European Conference on Computer Vision. pp. 1--17. Springer (2022)

\bibitem{antic2026lexis}
Anti{\'c}, D., Budria, A., Paschalidis, G., Dwivedi, S.K., Tzionas, D.: Lexis: Latent proximal interaction signatures for 3d hoi from an image. arXiv preprint arXiv:2604.20800  (2026)

\bibitem{banerjee2025hot3d}
Banerjee, P., Shkodrani, S., Moulon, P., Hampali, S., Han, S., Zhang, F., Zhang, L., Fountain, J., Miller, E., Basol, S., et~al.: Hot3d: Hand and object tracking in 3d from egocentric multi-view videos. In: Proceedings of the IEEE/CVF Conference on Computer Vision and Pattern Recognition. pp. 7061--7071 (2025)

\bibitem{barquero2025sparse}
Barquero, G., Bertsch, N., Marramreddy, M., Chac{\'o}n, C., Arcadu, F., Rigual, F., He, N.S., Palmero, C., Escalera, S., Ye, Y., et~al.: From sparse signal to smooth motion: Real-time motion generation with rolling prediction models. In: Proceedings of the Computer Vision and Pattern Recognition Conference. pp. 1850--1860 (2025)

\bibitem{ijcai2017-200}
Bhatnagar, B.L., Singh, S., Arora, C., Jawahar, C.: Unsupervised learning of deep feature representation for clustering egocentric actions. In: Proceedings of the Twenty-Sixth International Joint Conference on Artificial Intelligence, {IJCAI-17}. pp. 1447--1453 (2017)

\bibitem{bhatnagar2022behave}
Bhatnagar, B.L., Xie, X., Petrov, I.A., Sminchisescu, C., Theobalt, C., Pons-Moll, G.: Behave: Dataset and method for tracking human object interactions. In: Proceedings of the IEEE/CVF Conference on Computer Vision and Pattern Recognition. pp. 15935--15946 (2022)

\bibitem{braun2024physically}
Braun, J., Christen, S., Kocabas, M., Aksan, E., Hilliges, O.: Physically plausible full-body hand-object interaction synthesis. In: 2024 International Conference on 3D Vision (3DV). pp. 464--473. IEEE (2024)

\bibitem{cao2017egocentric}
Cao, C., Zhang, Y., Wu, Y., Lu, H., Cheng, J.: Egocentric gesture recognition using recurrent 3d convolutional neural networks with spatiotemporal transformer modules. 2017 IEEE International Conference on Computer Vision (ICCV)  (2017)

\bibitem{cao2025avatargo}
Cao, Y., Pan, L., Han, K., Wong, K.Y.K., Liu, Z.: Avatar{GO}: Zero-shot 4d human-object interaction generation and animation. In: The Thirteenth International Conference on Learning Representations (2025)

\bibitem{castillo2023bodiffusion}
Castillo, A., Escobar, M., Jeanneret, G., Pumarola, A., Arbel{\'a}ez, P., Thabet, A., Sanakoyeu, A.: Bodiffusion: Diffusing sparse observations for full-body human motion synthesis. In: Proceedings of the IEEE/CVF International Conference on Computer Vision. pp. 4221--4231 (2023)

\bibitem{chi2024estimating}
Chi, S., Huang, P.H., Sachdeva, E., Ma, H., Ramani, K., Lee, K.: Estimating ego-body pose from doubly sparse egocentric video data. Advances in Neural Information Processing Systems  \textbf{37},  55178--55203 (2024)

\bibitem{chiquier2023muscles}
Chiquier, M., Vondrick, C.: Muscles in action. In: Proceedings of the IEEE/CVF International Conference on Computer Vision. pp. 22091--22101 (2023)

\bibitem{christen2024diffh2o}
Christen, S., Hampali, S., Sener, F., Remelli, E., Hodan, T., Sauser, E., Ma, S., Tekin, B.: Diffh2o: Diffusion-based synthesis of hand-object interactions from textual descriptions. In: SIGGRAPH Asia 2024 Conference Papers. pp. 1--11 (2024)

\bibitem{dai2024hmd}
Dai, P., Zhang, Y., Liu, T., Fan, Z., Du, T., Su, Z., Zheng, X., Li, Z.: Hmd-poser: On-device real-time human motion tracking from scalable sparse observations. In: Proceedings of the IEEE/CVF Conference on Computer Vision and Pattern Recognition. pp. 874--884 (2024)

\bibitem{dai2024interfusion}
Dai, S., Li, W., Sun, H., Huang, H., Ma, C., Huang, H., Xu, K., Hu, R.: Interfusion: Text-driven generation of 3d human-object interaction. In: European Conference on Computer Vision. pp. 18--35. Springer (2024)

\bibitem{dai2022hsc4d}
Dai, Y., Lin, Y., Wen, C., Shen, S., Xu, L., Yu, J., Ma, Y., Wang, C.: Hsc4d: Human-centered 4d scene capture in large-scale indoor-outdoor space using wearable imus and lidar. In: Proceedings of the IEEE/CVF Conference on Computer Vision and Pattern Recognition. pp. 6792--6802 (2022)

\bibitem{dhariwal2021diffusion}
Dhariwal, P., Nichol, A.: Diffusion models beat gans on image synthesis. Advances in neural information processing systems  \textbf{34},  8780--8794 (2021)

\bibitem{diller2024cg}
Diller, C., Dai, A.: Cg-hoi: Contact-guided 3d human-object interaction generation. In: Proceedings of the IEEE/CVF Conference on Computer Vision and Pattern Recognition. pp. 19888--19901 (2024)

\bibitem{du2023avatars}
Du, Y., Kips, R., Pumarola, A., Starke, S., Thabet, A., Sanakoyeu, A.: Avatars grow legs: Generating smooth human motion from sparse tracking inputs with diffusion model. In: Proceedings of the IEEE/CVF Conference on Computer Vision and Pattern Recognition. pp. 481--490 (2023)

\bibitem{escobar2025egocast}
Escobar, M., Puentes, J., Forigua, C., Pont-Tuset, J., Maninis, K.K., Arbelaez, P.: Egocast: Forecasting egocentric human pose in the wild. In: 2025 IEEE/CVF Winter Conference on Applications of Computer Vision (WACV). pp. 5831--5841. IEEE (2025)

\bibitem{fan2023arctic}
Fan, Z., Taheri, O., Tzionas, D., Kocabas, M., Kaufmann, M., Black, M.J., Hilliges, O.: Arctic: A dataset for dexterous bimanual hand-object manipulation. In: Proceedings of the IEEE/CVF Conference on Computer Vision and Pattern Recognition. pp. 12943--12954 (2023)

\bibitem{fathi2011understanding}
Fathi, A., Farhadi, A., Rehg, J.M.: Understanding egocentric activities. In: 2011 international conference on computer vision. pp. 407--414. IEEE (2011)

\bibitem{feng2024stratified}
Feng, H., Ma, W., Gao, Q., Zheng, X., Xue, N., Xu, H.: Stratified avatar generation from sparse observations. In: Proceedings of the IEEE/CVF Conference on Computer Vision and Pattern Recognition. pp. 153--163 (2024)

\bibitem{fu2026egograsp}
Fu, H., Wang, W., Qiao, X., Yang, S., Liu, Z., Zhao, B.: Egograsp: World-space hand-object interaction estimation from egocentric videos. arXiv preprint arXiv:2601.01050  (2026)

\bibitem{ghosh2023imos}
Ghosh, A., Dabral, R., Golyanik, V., Theobalt, C., Slusallek, P.: Imos: Intent-driven full-body motion synthesis for human-object interactions. In: Computer Graphics Forum. vol.~42, pp. 1--12. Wiley Online Library (2023)

\bibitem{guzov24ireplica}
Guzov, V., Chibane, J., Marin, R., He, Y., Saracoglu, Y., Sattler, T., Pons-Moll, G.: Interaction replica: Tracking human-object interaction and scene changes from human motion. In: International Conference on 3D Vision (3DV) (March 2024)

\bibitem{guzov-jiang2025hmd2}
Guzov, V., Jiang, Y., Hong, F., Pons-Moll, G., Newcombe, R., Liu, C.K., Ye, Y., Ma, L.: Hmd$^2$: Environment-aware motion generation from single egocentric head-mounted device. In: International Conference on 3D Vision (3DV) (March 2025)

\bibitem{HPS}
Guzov, V., Mir, A., Sattler, T., Pons-Moll, G.: Human poseitioning system (hps): 3d human pose estimation and self-localization in large scenes from body-mounted sensors. In: {IEEE} Conference on Computer Vision and Pattern Recognition (CVPR). {IEEE} (jun 2021)

\bibitem{blendify2024}
Guzov, V., Petrov, I.A., Pons-Moll, G.: Blendify -- python rendering framework for blender. arXiv preprint arXiv:2410.17858  (2024)

\bibitem{hassan2019resolving}
Hassan, M., Choutas, V., Tzionas, D., Black, M.J.: Resolving 3d human pose ambiguities with 3d scene constraints. In: Proceedings of the IEEE/CVF international conference on computer vision. pp. 2282--2292 (2019)

\bibitem{hassan2021populating}
Hassan, M., Ghosh, P., Tesch, J., Tzionas, D., Black, M.J.: Populating 3d scenes by learning human-scene interaction. In: Proceedings of the IEEE/CVF Conference on Computer Vision and Pattern Recognition. pp. 14708--14718 (2021)

\bibitem{ho2020denoising}
Ho, J., Jain, A., Abbeel, P.: Denoising diffusion probabilistic models. Advances in neural information processing systems  \textbf{33},  6840--6851 (2020)

\bibitem{ho2022video}
Ho, J., Salimans, T., Gritsenko, A., Chan, W., Norouzi, M., Fleet, D.J.: Video diffusion models. Advances in Neural Information Processing Systems  \textbf{35},  8633--8646 (2022)

\bibitem{hollidt2024egosim}
Hollidt, D., Streli, P., Jiang, J., Haghighi, Y., Qian, C., Liu, X., Holz, C.: Egosim: An egocentric multi-view simulator and real dataset for body-worn cameras during motion and activity. Advances in Neural Information Processing Systems  \textbf{37},  106607--106627 (2024)

\bibitem{hong2025egolm}
Hong, F., Guzov, V., Kim, H.J., Ye, Y., Newcombe, R., Liu, Z., Ma, L.: Egolm: Multi-modal language model of egocentric motions. In: Proceedings of the Computer Vision and Pattern Recognition Conference. pp. 5344--5354 (2025)

\bibitem{huang2023diffusion}
Huang, S., Wang, Z., Li, P., Jia, B., Liu, T., Zhu, Y., Liang, W., Zhu, S.C.: Diffusion-based generation, optimization, and planning in 3d scenes. In: Proceedings of the IEEE/CVF Conference on Computer Vision and Pattern Recognition. pp. 16750--16761 (2023)

\bibitem{DIP:SIGGRAPHAsia:2018}
Huang, Y., Kaufmann, M., Aksan, E., Black, M.J., Hilliges, O., Pons-Moll, G.: Deep inertial poser: Learning to reconstruct human pose from sparse inertial measurements in real time. ACM Transactions on Graphics, (Proc. SIGGRAPH Asia)  \textbf{37}(6),  185:1--185:15 (nov 2018)

\bibitem{huang2024intercap}
Huang, Y., Taheri, O., Black, M.J., Tzionas, D.: {InterCap}: Joint markerless {3D} tracking of humans and objects in interaction from multi-view {RGB-D} images. {International Journal of Computer Vision (IJCV)}  (2024)

\bibitem{jiang2024manikin}
Jiang, J., Streli, P., Luo, X., Gebhardt, C., Holz, C.: Manikin: biomechanically accurate neural inverse kinematics for human motion estimation. In: European Conference on Computer Vision. pp. 128--146. Springer (2024)

\bibitem{jiang2024egoposer}
Jiang, J., Streli, P., Meier, M., Holz, C.: Egoposer: Robust real-time egocentric pose estimation from sparse and intermittent observations everywhere. In: European Conference on Computer Vision. pp. 277--294. Springer (2024)

\bibitem{jiang2022avatarposer}
Jiang, J., Streli, P., Qiu, H., Fender, A., Laich, L., Snape, P., Holz, C.: Avatarposer: Articulated full-body pose tracking from sparse motion sensing. In: European conference on computer vision. pp. 443--460. Springer (2022)

\bibitem{jiang2024scaling}
Jiang, N., Zhang, Z., Li, H., Ma, X., Wang, Z., Chen, Y., Liu, T., Zhu, Y., Huang, S.: Scaling up dynamic human-scene interaction modeling. In: Proceedings of the IEEE/CVF Conference on Computer Vision and Pattern Recognition. pp. 1737--1747 (2024)

\bibitem{jiang2022transformer}
Jiang, Y., Ye, Y., Gopinath, D., Won, J., Winkler, A.W., Liu, C.K.: Transformer inertial poser: Real-time human motion reconstruction from sparse imus with simultaneous terrain generation. In: SIGGRAPH Asia 2022 Conference Papers. pp.~1--9 (2022)

\bibitem{jiang2022neuralhofusion}
Jiang, Y., Jiang, S., Sun, G., Su, Z., Guo, K., Wu, M., Yu, J., Xu, L.: Neuralhofusion: Neural volumetric rendering under human-object interactions. In: Proceedings of the IEEE/CVF Conference on Computer Vision and Pattern Recognition. pp. 6155--6165 (2022)

\bibitem{kang2023ego3dpose}
Kang, T., Lee, K., Zhang, J., Lee, Y.: Ego3dpose: Capturing 3d cues from binocular egocentric views. In: SIGGRAPH Asia 2023 Conference Papers. pp. 1--10 (2023)

\bibitem{Kaufmann_Vechev_aitviewer_2022}
Kaufmann, M., Vechev, V., Mylonopoulos, D.: {aitviewer} (7 2022), \url{https://github.com/eth-ait/aitviewer}

\bibitem{kaufmann2021pose}
Kaufmann, M., Zhao, Y., Tang, C., Tao, L., Twigg, C., Song, J., Wang, R., Hilliges, O.: Em-pose: 3d human pose estimation from sparse electromagnetic trackers. In: Proceedings of the IEEE/CVF international conference on computer vision. pp. 11510--11520 (2021)

\bibitem{kister2026inhabit}
Kister, N., YM, P., S{\'a}r{\'a}ndi, I., Wang, J., Khoreva, A., Pons-Moll, G.: Inhabit: Leveraging image foundation models for scalable 3d human placement. arXiv preprint arXiv:2604.19673  (2026)

\bibitem{kulkarni2024nifty}
Kulkarni, N., Rempe, D., Genova, K., Kundu, A., Johnson, J., Fouhey, D., Guibas, L.: Nifty: Neural object interaction fields for guided human motion synthesis. In: Proceedings of the IEEE/CVF Conference on Computer Vision and Pattern Recognition. pp. 947--957 (2024)

\bibitem{lee2025rewind}
Lee, J., Xu, W., Richard, A., Wei, S.E., Saito, S., Bai, S., Wang, T.L., Sung, M., Kim, T.K., Saragih, J.: Rewind: Real-time egocentric whole-body motion diffusion with exemplar-based identity conditioning. In: Proceedings of the IEEE/CVF Conference on Computer Vision and Pattern Recognition. pp. 7095--7104 (2025)

\bibitem{lee2024mocap}
Lee, J., Joo, H.: Mocap everyone everywhere: Lightweight motion capture with smartwatches and a head-mounted camera. In: Proceedings of the IEEE/CVF conference on computer vision and pattern recognition. pp. 1091--1100 (2024)

\bibitem{li2025unimotion}
Li, C., Chibane, J., He, Y., Pearl, N., Geiger, A., Pons-Moll, G.: Unimotion: Unifying 3d human motion synthesis and understanding. In: 2025 International Conference on 3D Vision (3DV). pp. 240--249. IEEE (2025)

\bibitem{li2026frankenmotion}
Li, C., Xie, X., Cao, Y., Geiger, A., Pons-Moll, G.: {FrankenMotion}: Part-level human motion generation and composition. In: Proceedings of the IEEE/CVF Conference on Computer Vision and Pattern Recognition (CVPR) (2026)

\bibitem{li2024controllable}
Li, J., Clegg, A., Mottaghi, R., Wu, J., Puig, X., Liu, C.K.: Controllable human-object interaction synthesis. In: European Conference on Computer Vision. pp. 54--72. Springer (2024)

\bibitem{li2023ego}
Li, J., Liu, K., Wu, J.: Ego-body pose estimation via ego-head pose estimation. In: Proceedings of the IEEE/CVF Conference on Computer Vision and Pattern Recognition. pp. 17142--17151 (2023)

\bibitem{li2023object}
Li, J., Wu, J., Liu, C.K.: Object motion guided human motion synthesis. ACM Transactions on Graphics (TOG)  \textbf{42}(6),  1--11 (2023)

\bibitem{li2024task}
Li, Q., Wang, J., Loy, C.C., Dai, B.: Task-oriented human-object interactions generation with implicit neural representations. In: Proceedings of the IEEE/CVF Winter Conference on Applications of Computer Vision. pp. 3035--3044 (2024)

\bibitem{lin2026imu}
Lin, L., Xia, S., Lai, Z., Sun, L., Yang, J., Pei, L.: Imu-hoi: A symbiotic framework for coherent human-object interaction and motion capture via contact-conscious inertial fusion. In: Proceedings of the IEEE/CVF Conference on Computer Vision and Pattern Recognition. pp. 42901--42910 (2026)

\bibitem{liu2023egofish3d}
Liu, Y., Yang, J., Gu, X., Chen, Y., Guo, Y., Yang, G.Z.: Egofish3d: Egocentric 3d pose estimation from a fisheye camera via self-supervised learning. IEEE Transactions on Multimedia  \textbf{25},  8880--8891 (2023)

\bibitem{liu2023egohmr}
Liu, Y., Yang, J., Gu, X., Guo, Y., Yang, G.Z.: Egohmr: Egocentric human mesh recovery via hierarchical latent diffusion model. In: 2023 IEEE International Conference on Robotics and Automation (ICRA). pp. 9807--9813. IEEE (2023)

\bibitem{loshchilov2017decoupled}
Loshchilov, I., Hutter, F.: Decoupled weight decay regularization. arXiv preprint arXiv:1711.05101  (2017)

\bibitem{ma2024nymeria}
Ma, L., Ye, Y., Hong, F., Guzov, V., Jiang, Y., Postyeni, R., Pesqueira, L., Gamino, A., Baiyya, V., Kim, H.J., et~al.: Nymeria: A massive collection of multimodal egocentric daily motion in the wild. In: European Conference on Computer Vision. pp. 445--465. Springer (2024)

\bibitem{ma2016going}
Ma, M., Fan, H., Kitani, K.M.: Going deeper into first-person activity recognition. In: Proceedings of the IEEE Conference on Computer Vision and Pattern Recognition. pp. 1894--1903 (2016)

\bibitem{AMASS:ICCV:2019}
Mahmood, N., Ghorbani, N., Troje, N.F., Pons-Moll, G., Black, M.J.: {AMASS}: Archive of motion capture as surface shapes. In: International Conference on Computer Vision. pp. 5442--5451 (Oct 2019)

\bibitem{de2025monado}
de~Mayo, M., Cremers, D., Pire, T.: The monado slam dataset for egocentric visual-inertial tracking. In: 2025 IEEE/RSJ International Conference on Intelligent Robots and Systems (IROS). pp. 13111--13118. IEEE (2025)

\bibitem{MetaEMG}
Meta EMG Wearable Technology (accessed February 7, 2026), \href{https://www.meta.com/en-gb/emerging-tech/emg-wearable-technology}{https://www.meta.com/en-gb/emerging-tech/emg-wearable-technology}

\bibitem{mir2024generating}
Mir, A., Puig, X., Kanazawa, A., Pons-Moll, G.: Generating continual human motion in diverse 3d scenes. In: 2024 International Conference on 3D Vision (3DV). pp. 903--913. IEEE (2024)

\bibitem{mollyn2023imuposer}
Mollyn, V., Arakawa, R., Goel, M., Harrison, C., Ahuja, K.: Imuposer: Full-body pose estimation using imus in phones, watches, and earbuds. In: Proceedings of the 2023 CHI Conference on Human Factors in Computing Systems. pp. 1--12 (2023)

\bibitem{nam2024joint}
Nam, H., Jung, D.S., Moon, G., Lee, K.M.: Joint reconstruction of 3d human and object via contact-based refinement transformer. In: Proceedings of the IEEE/CVF Conference on Computer Vision and Pattern Recognition. pp. 10218--10227 (2024)

\bibitem{nazarenus2026actionplan}
Nazarenus, E., Li, C., He, Y., Xie, X., Lenssen, J.E., Pons-Moll, G.: Actionplan: Future-aware streaming motion synthesis via frame-level action planning. In: European Conference on Computer Vision. Springer (2026)

\bibitem{OpenRing}
OpenRing: world's first Open-source Smart Ring (accessed February 7, 2026), \href{https://o-ring.tech/}{https://o-ring.tech/}

\bibitem{pan2025lookout}
Pan, B., Harley, A.W., Engelmann, F., Liu, C.K., Guibas, L.J.: Lookout: Real-world humanoid egocentric navigation. In: Proceedings of the IEEE/CVF International Conference on Computer Vision. pp. 24977--24988 (2025)

\bibitem{pan2023aria}
Pan, X., Charron, N., Yang, Y., Peters, S., Whelan, T., Kong, C., Parkhi, O., Newcombe, R., Ren, Y.C.: Aria digital twin: A new benchmark dataset for egocentric 3d machine perception. In: Proceedings of the IEEE/CVF International Conference on Computer Vision. pp. 20133--20143 (2023)

\bibitem{patel2025uniegomotion}
Patel, C., Nakamura, H., Kyuragi, Y., Kozuka, K., Niebles, J.C., Adeli, E.: Uniegomotion: A unified model for egocentric motion reconstruction, forecasting, and generation. In: Proceedings of the IEEE/CVF International Conference on Computer Vision. pp. 10318--10329 (2025)

\bibitem{SMPL-X:2019}
Pavlakos, G., Choutas, V., Ghorbani, N., Bolkart, T., Osman, A.A.A., Tzionas, D., Black, M.J.: Expressive body capture: {3D} hands, face, and body from a single image. In: Proceedings IEEE Conf. on Computer Vision and Pattern Recognition (CVPR). pp. 10975--10985 (2019)

\bibitem{peebles2023scalable}
Peebles, W., Xie, S.: Scalable diffusion models with transformers. In: Proceedings of the IEEE/CVF international conference on computer vision. pp. 4195--4205 (2023)

\bibitem{peng2023hoi}
Peng, X., Xie, Y., Wu, Z., Jampani, V., Sun, D., Jiang, H.: Hoi-diff: Text-driven synthesis of 3d human-object interactions using diffusion models. In: Proceedings of the Computer Vision and Pattern Recognition Conference. pp. 2878--2888 (2025)

\bibitem{petrov2023popup}
Petrov, I.A., Marin, R., Chibane, J., Pons-Moll, G.: Object pop-up: Can we infer 3d objects and their poses from human interactions alone? In: Proceedings of the IEEE/CVF Conference on Computer Vision and Pattern Recognition (2023)

\bibitem{petrov2024tridi}
Petrov, I.A., Marin, R., Chibane, J., Pons-Moll, G.: Tridi: Trilateral diffusion of 3d humans, objects, and interactions. In: Proceedings of the IEEE/CVF International Conference on Computer Vision (2025)

\bibitem{Aria}
Project Aria (accessed February 7, 2026), \href{https://www.projectaria.com/glasses/}{https://www.projectaria.com/glasses/}

\bibitem{qian2022pointnext}
Qian, G., Li, Y., Peng, H., Mai, J., Hammoud, H., Elhoseiny, M., Ghanem, B.: Pointnext: Revisiting pointnet++ with improved training and scaling strategies. Advances in neural information processing systems  \textbf{35},  23192--23204 (2022)

\bibitem{ramesh2022hierarchical}
Ramesh, A., Dhariwal, P., Nichol, A., Chu, C., Chen, M.: Hierarchical text-conditional image generation with clip latents. arXiv preprint arXiv:2204.06125  \textbf{1}(2), ~3 (2022)

\bibitem{rempe2021humor}
Rempe, D., Birdal, T., Hertzmann, A., Yang, J., Sridhar, S., Guibas, L.J.: Humor: 3d human motion model for robust pose estimation. In: Proceedings of the IEEE/CVF international conference on computer vision. pp. 11488--11499 (2021)

\bibitem{rhodin2016egocap}
Rhodin, H., Richardt, C., Casas, D., Insafutdinov, E., Shafiei, M., Seidel, H.P., Schiele, B., Theobalt, C.: Egocap: egocentric marker-less motion capture with two fisheye cameras. ACM Transactions on Graphics (TOG)  \textbf{35}(6),  1--11 (2016)

\bibitem{rogez2015first}
Rogez, G., Supancic, J.S., Ramanan, D.: First-person pose recognition using egocentric workspaces. In: Proceedings of the IEEE conference on computer vision and pattern recognition. pp. 4325--4333 (2015)

\bibitem{MANO:SIGGRAPHASIA:2017}
Romero, J., Tzionas, D., Black, M.J.: Embodied hands: Modeling and capturing hands and bodies together. ACM Transactions on Graphics, (Proc. SIGGRAPH Asia)  \textbf{36}(6) (Nov 2017)

\bibitem{shin2026egomdm}
Shin, S., Pahuja, A., Richard, A., Kitani, K., Saragih, J., Chen, Y., Xu, W., Halilaj, E., Bagautdinov, T.: Egomdm: Diffusion-based human motion synthesis from sparse egocentric sensors. In: Thirteenth International Conference on 3D Vision (2026)

\bibitem{sohl2015deep}
Sohl-Dickstein, J., Weiss, E., Maheswaranathan, N., Ganguli, S.: Deep unsupervised learning using nonequilibrium thermodynamics. In: International conference on machine learning. pp. 2256--2265. PMLR (2015)

\bibitem{song2024hoianimator}
Song, W., Zhang, X., Li, S., Gao, Y., Hao, A., Hou, X., Chen, C., Li, N., Qin, H.: Hoianimator: Generating text-prompt human-object animations using novel perceptive diffusion models. In: Proceedings of the IEEE/CVF Conference on Computer Vision and Pattern Recognition. pp. 811--820 (2024)

\bibitem{su2024roformer}
Su, J., Ahmed, M., Lu, Y., Pan, S., Bo, W., Liu, Y.: Roformer: Enhanced transformer with rotary position embedding. Neurocomputing  \textbf{568},  127063 (2024)

\bibitem{sarandi24nlf}
Sárándi, I., Pons-Moll, G.: Neural localizer fields for continuous 3d human pose and shape estimation. Advances in Neural Information Processing Systems (NeurIPS)  (2024)

\bibitem{tang2024unified}
Tang, J., Wang, J., Ji, K., Xu, L., Yu, J., Shi, Y.: A unified diffusion framework for scene-aware human motion estimation from sparse signals. In: Proceedings of the IEEE/CVF Conference on Computer Vision and Pattern Recognition. pp. 21251--21262 (2024)

\bibitem{tevet2022human}
Tevet, G., Raab, S., Gordon, B., Shafir, Y., Cohen-Or, D., Bermano, A.H.: Human motion diffusion model. In: The Eleventh International Conference on Learning Representations (2023)

\bibitem{tome2020selfpose}
Tome, D., Alldieck, T., Peluse, P., Pons-Moll, G., Agapito, L., Badino, H., De~la Torre, F.: Selfpose: 3d egocentric pose estimation from a headset mounted camera. IEEE Transactions on Pattern Analysis and Machine Intelligence  \textbf{45}(6),  6794--6806 (2020)

\bibitem{tripathi2023deco}
Tripathi, S., Chatterjee, A., Passy, J.C., Yi, H., Tzionas, D., Black, M.J.: Deco: Dense estimation of 3d human-scene contact in the wild. In: Proceedings of the IEEE/CVF International Conference on Computer Vision. pp. 8001--8013 (2023)

\bibitem{vlasic2007practical}
Vlasic, D., Adelsberger, R., Vannucci, G., Barnwell, J., Gross, M., Matusik, W., Popovi{\'c}, J.: Practical motion capture in everyday surroundings. ACM transactions on graphics (TOG)  \textbf{26}(3),  35--es (2007)

\bibitem{von2017sparse}
Von~Marcard, T., Rosenhahn, B., Black, M.J., Pons-Moll, G.: Sparse inertial poser: Automatic 3d human pose estimation from sparse imus. In: Computer graphics forum. vol.~36, pp. 349--360. Wiley Online Library (2017)

\bibitem{wang2024egocentric}
Wang, J., Cao, Z., Luvizon, D., Liu, L., Sarkar, K., Tang, D., Beeler, T., Theobalt, C.: Egocentric whole-body motion capture with fisheyevit and diffusion-based motion refinement. In: Proceedings of the IEEE/CVF Conference on Computer Vision and Pattern Recognition. pp. 777--787 (2024)

\bibitem{winkler2022questsim}
Winkler, A., Won, J., Ye, Y.: Questsim: Human motion tracking from sparse sensors with simulated avatars. In: SIGGRAPH Asia 2022 Conference Papers. pp.~1--8 (2022)

\bibitem{xia2025envposer}
Xia, S., Zhang, Y., Su, Z., Zheng, X., Lv, Z., Wang, G., Zhang, Y., Wu, Q., Chu, L., Pei, L.: Envposer: Environment-aware realistic human motion estimation from sparse observations with uncertainty modeling. In: Proceedings of the Computer Vision and Pattern Recognition Conference. pp. 1839--1849 (2025)

\bibitem{xie2024template}
Xie, X., Bhatnagar, B.L., Lenssen, J.E., Pons-Moll, G.: Template free reconstruction of human-object interaction with procedural interaction generation. In: Proceedings of the IEEE/CVF Conference on Computer Vision and Pattern Recognition. pp. 10003--10015 (2024)

\bibitem{xie2022chore}
Xie, X., Bhatnagar, B.L., Pons-Moll, G.: Chore: Contact, human and object reconstruction from a single rgb image. In: European Conference on Computer Vision ({ECCV}). {Springer} (October 2022)

\bibitem{xie2023vistracker}
Xie, X., Bhatnagar, B.L., Pons-Moll, G.: Visibility aware human-object interaction tracking from single rgb camera. In: IEEE Conference on Computer Vision and Pattern Recognition (CVPR) (June 2023)

\bibitem{xie2024InterTrack}
Xie, X., Lenssen, J.E., Pons-Moll, G.: Intertrack: Tracking human object interaction without object templates. In: International Conference on 3D Vision 2025 (2025)

\bibitem{xu2024regennet}
Xu, L., Zhou, Y., Yan, Y., Jin, X., Zhu, W., Rao, F., Yang, X., Zeng, W.: Regennet: Towards human action-reaction synthesis. In: Proceedings of the IEEE/CVF Conference on Computer Vision and Pattern Recognition. pp. 1759--1769 (2024)

\bibitem{xu2023interdiff}
Xu, S., Li, Z., Wang, Y.X., Gui, L.Y.: Interdiff: Generating 3d human-object interactions with physics-informed diffusion. In: Proceedings of the IEEE/CVF International Conference on Computer Vision. pp. 14928--14940 (2023)

\bibitem{xu2026interprior}
Xu, S., Schulter, S., Ziyadi, M., He, X., Fei, X., Wang, Y.X., Gui, L.Y.: Interprior: Scaling generative control for physics-based human-object interactions. In: Proceedings of the IEEE/CVF Conference on Computer Vision and Pattern Recognition. pp. 23516--23527 (2026)

\bibitem{xu2024interdreamer}
Xu, S., Wang, Y.X., Gui, L., et~al.: Interdreamer: Zero-shot text to 3d dynamic human-object interaction. Advances in Neural Information Processing Systems  \textbf{37},  52858--52890 (2024)

\bibitem{xu2024mobileposer}
Xu, V., Gao, C., Hoffmann, H., Ahuja, K.: Mobileposer: Real-time full-body pose estimation and 3d human translation from imus in mobile consumer devices. In: Proceedings of the 37th Annual ACM Symposium on User Interface Software and Technology. pp. 1--11 (2024)

\bibitem{xu2019mo2cap2}
Xu, W., Chatterjee, A., Zollhoefer, M., Rhodin, H., Fua, P., Seidel, H.P., Theobalt, C.: {Mo}$^{2}${Cap}$^{2}$ : Real-time mobile 3d motion capture with a cap-mounted fisheye camera. IEEE Transactions on Visualization and Computer Graphics pp.~1--1 (2019)

\bibitem{ym2025physic}
Yalandur~Muralidhar, P., Xue, Y., Xie, X., Kostyrko, M., Pons-Moll, G.: Physic: Physically plausible 3d human-scene interaction and contact from a single image (2025)

\bibitem{yang2024f}
Yang, J., Niu, X., Jiang, N., Zhang, R., Huang, S.: F-hoi: Toward fine-grained semantic-aligned 3d human-object interactions. In: European Conference on Computer Vision. pp. 91--110. Springer (2024)

\bibitem{yang2024egochoir}
Yang, Y., Zhai, W., Wang, C., Yu, C., Cao, Y., Zha, Z.J.: Egochoir: Capturing 3d human-object interaction regions from egocentric views. Advances in Neural Information Processing Systems  \textbf{37},  54529--54557 (2024)

\bibitem{ye2026whole}
Ye, Y., Li, J., Rong, R., Liu, C.K.: Whole: World-grounded hand-object lifted from egocentric videos. arXiv preprint arXiv:2602.22209  (2026)

\bibitem{yi2025egoallo}
Yi, B., Ye, V., Zheng, M., Li, Y., M{\"u}ller, L., Pavlakos, G., Ma, Y., Malik, J., Kanazawa, A.: Estimating body and hand motion in an ego-sensed world. In: Proceedings of the Computer Vision and Pattern Recognition Conference. pp. 7072--7084 (2025)

\bibitem{yi2022physical}
Yi, X., Zhou, Y., Habermann, M., Shimada, S., Golyanik, V., Theobalt, C., Xu, F.: Physical inertial poser (pip): Physics-aware real-time human motion tracking from sparse inertial sensors. In: Proceedings of the IEEE/CVF conference on computer vision and pattern recognition. pp. 13167--13178 (2022)

\bibitem{yi2021transpose}
Yi, X., Zhou, Y., Xu, F.: Transpose: Real-time 3d human translation and pose estimation with six inertial sensors. ACM Transactions On Graphics (TOG)  \textbf{40}(4),  1--13 (2021)

\bibitem{liu2024egohdm}
Yin, H., Liu, B., Kaufmann, M., He, J., Christen, S., Song, J., Hui, P.: Egohdm: A real-time egocentric-inertial human motion capture, localization, and dense mapping system. ACM Transactions on Graphics (TOG)  \textbf{43}(6),  1--12 (2024)

\bibitem{ym2026graft}
YM, P., Xue, Y., Chen, Y., Kister, N., S{\'a}r{\'a}ndi, I., Pons-Moll, G.: Graft: Geometric refinement and fitting transformer for human scene reconstruction (2026)

\bibitem{yonemoto2015egocentric}
Yonemoto, H., Murasaki, K., Osawa, T., Sudo, K., Shimamura, J., Taniguchi, Y.: Egocentric articulated pose tracking for action recognition. In: International Conference on Machine Vision Applications (MVA) (2015)

\bibitem{yoon2026egoxtreme}
Yoon, T., Han, Y., Ji, S., Park, J., Kim, S., Kwon, T., Kim, H.S.: Egoxtreme: A dataset for robust object pose estimation in egocentric views under extreme conditions. In: Proceedings of the IEEE/CVF Conference on Computer Vision and Pattern Recognition. pp. 40696--40706 (2026)

\bibitem{zhang2023neuraldome}
Zhang, J., Luo, H., Yang, H., Xu, X., Wu, Q., Shi, Y., Yu, J., Xu, L., Wang, J.: Neuraldome: A neural modeling pipeline on multi-view human-object interactions. In: Proceedings of the IEEE/CVF Conference on Computer Vision and Pattern Recognition. pp. 8834--8845 (2023)

\bibitem{zhang2024hoi}
Zhang, J., Zhang, J., Song, Z., Shi, Z., Zhao, C., Shi, Y., Yu, J., Xu, L., Wang, J.: Hoi-m\^{} 3: Capture multiple humans and objects interaction within contextual environment. In: Proceedings of the IEEE/CVF Conference on Computer Vision and Pattern Recognition. pp. 516--526 (2024)

\bibitem{zhang2022egobody}
Zhang, S., Ma, Q., Zhang, Y., Qian, Z., Kwon, T., Pollefeys, M., Bogo, F., Tang, S.: Egobody: Human body shape and motion of interacting people from head-mounted devices. In: European conference on computer vision. pp. 180--200. Springer (2022)

\bibitem{zhang2024force}
Zhang, X., Bhatnagar, B.L., Starke, S., Petrov, I., Guzov, V., Dhamo, H., P{\'e}rez-Pellitero, E., Pons-Moll, G.: Force: Physics-aware human-object interaction. In: 2025 International Conference on 3D Vision (3DV). pp. 1473--1486. IEEE (2025)

\bibitem{zhang2024scenic}
Zhang, X., Starke, S., Guzov, V., Zhang, Z., Pellitero, E.P., Pons-Moll, G.: Scenic: Scene-aware semantic navigation with instruction-guided control. arXiv preprint arXiv:2412.15664  (2024)

\bibitem{zhang2026glove2hand}
Zhang, X., Kou, Z., Qin, C., Huang, M., Ristani, E., Kumar, A., Chen, L., He, K., Boularias, A., Guan, L.: Glove2hand: Synthesizing natural hand-object interaction from multi-modal sensing gloves. In: Proceedings of the IEEE/CVF Conference on Computer Vision and Pattern Recognition. pp. 1829--1840 (2026)

\bibitem{zhao2024imhoi}
Zhao, C., Zhang, J., Du, J., Shan, Z., Wang, J., Yu, J., Wang, J., Xu, L.: I'm hoi: Inertia-aware monocular capture of 3d human-object interactions. In: Proceedings of the IEEE/CVF Conference on Computer Vision and Pattern Recognition. pp. 729--741 (2024)

\bibitem{zhao2024instance}
Zhao, Y., Ma, H., Kong, S., Fowlkes, C.: Instance tracking in 3d scenes from egocentric videos. In: Proceedings of the IEEE/CVF Conference on Computer Vision and Pattern Recognition. pp. 21933--21944 (2024)

\bibitem{zheng2023realistic}
Zheng, X., Su, Z., Wen, C., Xue, Z., Jin, X.: Realistic full-body tracking from sparse observations via joint-level modeling. In: Proceedings of the IEEE/CVF International Conference on Computer Vision. pp. 14678--14688 (2023)

\bibitem{zhou2019continuity}
Zhou, Y., Barnes, C., Lu, J., Yang, J., Li, H.: On the continuity of rotation representations in neural networks. In: Proceedings of the IEEE/CVF conference on computer vision and pattern recognition. pp. 5745--5753 (2019)

\bibitem{zuo2024loose}
Zuo, C., Wang, Y., Zhan, L., Guo, S., Yi, X., Xu, F., Qin, Y.: Loose inertial poser: Motion capture with imu-attached loose-wear jacket. In: Proceedings of the IEEE/CVF Conference on Computer Vision and Pattern Recognition. pp. 2209--2219 (2024)

\end{thebibliography}
}

\clearpage

\begin{center}
    {\Large \bf Supplementary Material for \\ \vspace{2pt} ECHO: Ego-Centric modeling of Human-Object interactions}
    \vspace{20pt}
\end{center}

\begin{quotation}
\noindent \textbf{Abstract.}
This supplementary material provides a discussion on the broader impacts of our work in \cref{sec:a_broader}. In \cref{sec:a_background}, we provide background on the diffusion process and a summary of the notation used in the text. We present additional experimental results, including evaluations with contact conditioning, ablation studies on the OMOMO dataset in \cref{sec:a_eval}. We report further implementation details, including the training objective, inference guidance, and smooth inpainting formulation in \cref{sec:a_implementation}. Finally, in \cref{sec:a_metrics}, we provide full definitions of the evaluation metrics. 
In the supplementary video, we show dynamic visualizations of the generated results.
\end{quotation}

\setcounter{figure}{0}
\setcounter{table}{0}
\renewcommand{\thefigure}{S\arabic{figure}}
\renewcommand{\thetable}{S\arabic{table}}

\section{Broader impacts}
\label{sec:a_broader}
The ability of our model to capture and generate continuous human-object interactions offers significant value for fields such as digital content creation and ergonomics. 
This research direction could enable new applications for studying human behavior and developing realistic virtual experiences.
However, this technology can also be misused. 
Tracking detailed human actions could lead to unauthorized surveillance, creating privacy issues. 
We recognize that future advances might make this technology easier to abuse. 
Therefore, we believe that its responsible development must be an ongoing priority.

\section{Background and Notation}
\renewcommand{\arraystretch}{1.2}
\begin{table}[t!]
    \centering
    \small
    \setlength{\tabcolsep}{3pt}
    \caption{\label{tab:sym_table}\textbf{Notation Table}. The main notation used in our paper.}
    \begin{tabular}{lll}
    Symbol               & Description                          & Domain \\
    \hline
    $ \window $          & Network's temporal window            & $60$ frames \\
    $ \sequence $        & Input sequence length                & $\mathbb{N}$ \\
    $ \set{T}_\h, \set{T}_\obj, \set{T}_\inter $        & Denoising step for each modality                & $\{0, 1, \dots, 1000\}$ \\
    \hline
    $ \condego $         & 3 point egocentric conditioning      & $\left\{
                                                                  \begin{array}{l}
                                                                  [\egoheaddeltat, \egorotcan,\\
                                                                  \egoheight, \egohanddeltat, \egorothandscan]^{1..\sequence}
                                                                  \end{array}
                                                                  \right\}$ \\
    $ \egorotcan $      & Canonicalized head rotation           & Formal - $\mathbb{R}^{3\times3}$ / Network - $\mathbb{R}^{6}$ \\
    $ \egoheaddeltat $  & Relative head SE3 transformation      & Formal - $\mathbb{R}^{4\times4}$ / Network - $\mathbb{R}^{9}$ \\
    $ \egoheight $      & Head to floor distance                & $\mathbb{R}^{1}$ \\
    $ \egorothandscan $ & Canonicalized hands rotation          & Formal - $\mathbb{R}^{2\times3\times3}$ / Network - $\mathbb{R}^{2\times6}$ \\
    $ \egohanddeltat $  & Relative hands SE3 transformation     & Formal - $\mathbb{R}^{2\times4\times4}$ / Network - $\mathbb{R}^{2\times9}$ \\
    \hline 
    $ \h $               & Human Modality                       & $\{ [\hpose]^{1..\sequence} \}$ \\
    $ \hpose $           & Human Pose                           & Formal - $\mathbb{R}^{21 \times 3}$ / Network - $\mathbb{R}^{21\times6}$ \\
    $ \hshape $          & Human Shape parameters               & $\mathbb{R}^{10}$ \\
    $ \htemplatev $      & Human Template's Vertices            & $\mathbb{R}^{10475}$ \\
    $ \hglobal $         & Human Global SE3 transformation      & Formal - $\mathbb{R}
    ^{4\times4}$ \\
    $ \hjoints $         & Human Joints                         & $\mathbb{R}^{21\times3}$ \\
    $ \hvelocity $       & Linear velocity of Human Joints      & $\mathbb{R}^{21\times3}$ \\
    \hline
    $ \obj $             & Object Modality                      & $\{ \objglobal^{1..\sequence} \}$ \\
    $ \objglobal $       & Object Global SE3 transformation     & Formal - $\mathbb{R}^{4\times4}$ / Network - $\mathbb{R}^{9}$ \\
    $ \condobj $         & Object Information for conditioning  & $(\objfeatures, \objonehot)$ \\
    $ \objfeatures $     & PointNext object features            & $\mathbb{R}^{1024}$ \\
    $ \objonehot $       & one-hot encoding of the class        & $\{0,1\}^{34}$ \\
    
    $ \objtemplatev $    &   Object Template's Vertices         & $\mathbb{R}^{1500}$ \\
    \hline
    $ \inter $           & Interaction                          & $\inter=\{ \intervec^{1..\sequence} \}$ \\
    $ \intervec $        & Vector of contact labels             & $\{\intervec^{\text{HOI}}, \intervec^{\text{Env}}\}$ \\
    $ \intervec^{\text{HOI}} $ & Human-object contacts & $[0, 1]^{64}$ \\
    $ \intervec^{\text{Env}} $ & Human-floor contacts & $[0, 1]^{8}$ \\
    $ \interpoints $     & Contact points on the human body     & $\mathbb{R}^{64}$, $\interpoints \subset \htemplatev$ \\
    $ \vec{d} $ & Distances between $\interpoints$ and $ \objtemplatev $ & $\mathbb{R}^{64}$ \\
    \hline
    
    \end{tabular}
\end{table}

\renewcommand{\arraystretch}{1.0}
\label{sec:a_background}
\subsubsection{Background.} 
The forward diffusion process can be formulated as a Markov chain with $T$ steps. 
Starting from a clean sample $\mathbf{z}^0$, it produces a series of distributions $q(\mathbf{z}^\set{T} | \mathbf{z}^{\set{T}-1})$: $q(\mathbf{z}^{1:T} | \mathbf{z}^0) = \prod_{\set{T}=1}^{T}{q(\mathbf{z}^{\set{T}} | \mathbf{z}^{\set{T}-1})}$.
We add noise to the distribution for $T$ steps, until finally $\mathbf{z}^\set{T}$ becomes a sample from $\mathcal{N}(\mathbf{0},\mathbf{I})$.
Defining $\beta_0=0$, and $\beta_\set{T} \in (0,1)$ we obtain:
\begin{equation}
    \begin{aligned}
        q(\mathbf{z}^\set{T} | \mathbf{z}^{\set{T}-1}) & = \mathcal{N}(
          \mathbf{z}^\set{T}; 
          \sqrt{1 - \beta_\set{T}} \mathbf{z}^{\set{T}-1},
          \beta_\set{T} \mathbf{I}
        ).
    \end{aligned}
\end{equation}

Formulation of DDPM~\cite{ho2020denoising} allows us to obtain a closed-form expression for $\mathbf{z}^\set{T}$. 
Let $\alpha_i=1 - \beta_i$, $\bar{\alpha}_\set{T}=\prod_{i=1}^{\set{T}}{\alpha_i}$, and $\epsilon \sim \mathcal{N}(0, \mathbf{I})$:
\begin{equation}
    \begin{aligned}
        q(\mathbf{z}^\set{T} | \mathbf{z}^0) & = \mathcal{N}(
          \mathbf{z}^\set{T};
          \sqrt{\bar{\alpha}_\set{T}}\mathbf{z}^0,
          (1 - \bar{\alpha}_\set{T}) \mathbf{I}
        ), \\
        \mathbf{z}^\set{T} & = \sqrt{\bar{\alpha}_\set{T}} \mathbf{z}^0 + 
          \sqrt{1 - \bar{\alpha}_\set{T}} \epsilon.
    \label{eqn:sup_ddpm}
    \end{aligned}
\end{equation}

Reversing the process, we obtain a formulation for the inference. 
Concretely, starting from $\mathbf{z}^T \sim \mathcal{N}(0, \mathbf{I})$, we can step-by-step recover the sample from the original distribution. 
We train our network to recover the original sample $\mathbf{z}^0$ directly as in~\cite{ramesh2022hierarchical} (instead of the traditional formulation, in which the added noise $\epsilon$ is recovered).
To achieve this, we parametrize the reverse process by a denoising neural network $\mathcal{D}_\psi$ that is trained to recover the original sample $\mathbf{z}^0$ from the noised sample $\mathbf{z}^\set{T}$ at denoising step $\set{T}$ given the condition $c$. Defining for brevity 
$\mathbb{E}_p \equiv \mathbb{E}_{\mathbf{z}^0 \sim p_{data}}$, 
$\mathbb{E}_\set{T} \equiv \mathbb{E}_{\set{T} \sim \mathcal{U}\{0,...,T\}}$, and 
$\mathbb{E}_q \equiv \mathbb{E}_{\mathbf{z}^\set{T} \sim q(\mathbf{z}^\set{T} | \mathbf{z}^0)}$ 
we obtain the training objective:
\begin{equation}
    \begin{aligned}
        \min_\psi  
           \mathbb{E}_p\, \mathbb{E}_\set{T}\, \mathbb{E}_q\, \|\mathcal{D}_\psi(\mathbf{z}^\set{T}; c, \set{T}) - \mathbf{z}^0\|_2.
    \end{aligned}
    \label{eqn:sup_dif_objective}
\end{equation}

An iterative denoising process with denoising network $\mathcal{D}_\psi$ is defined by the following:

\begin{equation}
    \mathbf{z}^{\set{T}-1} = \sqrt{\bar{\alpha}_{\set{T} - 1}} 
      \mathcal{D}_\psi(\mathbf{z}^\set{T}; c, \set{T}) + 
      \sqrt{1 - \bar{\alpha}_{\set{T}-1}} \epsilon,
    \label{eqn:diffusion_inference}
\end{equation}

where $\hat{\mathbf{z}}^0 = \mathcal{D}_\psi(\mathbf{z}^\set{T}; c, \set{T})$.

\subsubsection{Notation.} Tab.~\ref{tab:sym_table} defines symbols used in our work.

\section{Additional evaluation}
\label{sec:a_eval}

\renewcommand{\arraystretch}{1.2}
\begin{table}[h!]
    \setlength{\tabcolsep}{1.6pt}
    \centering
    \caption{\textbf{Fine-tuning on AMASS}. A two-stage variant ($\mathsf{A\text{-}FT}$) degrades substantially, confirming the gain comes from our tri-variate formulation, not data access.}

    \begin{tabular}{lccccc}
    \specialrule{.1em}{.05em}{.05em}
        
    \multirow{2}{*}{\bf Method} & \textbf{AMASS} & \multicolumn{2}{c}{\textbf{BEHAVE}} & \multicolumn{2}{c}{\textbf{OMOMO}} \\
        \cmidrule(lr){2-2} \cmidrule(lr){3-4} \cmidrule(lr){5-6}
         & MPJPE$\downarrow$ & MPJPE$\downarrow$ & $E_{v2v}\downarrow$ & MPJPE$\downarrow$ & $E_{v2v}\downarrow$ \\
        \cmidrule(lr){2-2} \cmidrule(lr){3-4} \cmidrule(lr){5-6}
        {\scriptsize \bodiffo}       &
             ${11.4^{\pm{0.3}}}$ & ${8.3^{\pm{0.2}}}$ & ${44.2^{\pm{1.2}}}$ & ${7.6^{\pm{0.4}}}$ & ${33.2^{\pm{1.9}}}$ \\
        {\scriptsize \egoalloho}    &
             $\second{8.9^{\pm{0.1}}}$ & $\second{7.6^{\pm{0.1}}}$ & ${39.1^{\pm{1.1}}}$ & $\second{6.6^{\pm{0.1}}}$ & ${30.8^{\pm{0.9}}}$ \\
        \hline
        {\scriptsize \methodname~$\mathsf{A\text{-}FT}$}    & 
            ${15.9^{\pm{0.7}}}$ & ${8.6^{\pm{0.2}}}$ & $\second{34.1^{\pm{1.6}}}$ & ${7.5^{\pm{0.2}}}$ & $\second{26.9^{\pm{1.9}}}$ \\
        \hline
        {\scriptsize \methodname \textbf{(Ours)}}             & 
            $\first{7.4^{\pm{0.1}}}$ & $\first{6.8^{\pm{0.1}}}$ & $\first{33.5^{\pm{0.5}}}$ & $\first{6.0^{\pm{0.1}}}$ & $\first{26.5^{\pm{1.1}}}$ \\
    \specialrule{.1em}{.05em}{.05em} 
    \end{tabular}

    \label{tab:sup_amass_ft}    
\end{table}
\renewcommand{\arraystretch}{1.0}

\renewcommand{\arraystretch}{1.2}
\begin{table}[h!]
    \setlength{\tabcolsep}{3pt}
    \centering
    \caption{\textbf{Physical plausibility analysis}. {\methodname} stays close to GT and balances contact and penetration, while baselines produce floaters and penetrations.}

    \begin{tabular}{lcccc}
        \specialrule{.1em}{.05em}{.05em} 
        \multicolumn{5}{c}{\textbf{BEHAVE}}\\
        \cmidrule{2-5}
                \bf Method & Min. D.$\downarrow$ & $C_{\%}\uparrow$ & $PD\downarrow$ & $P_{sc}\downarrow$ \\
         \cmidrule{2-5}
        Data             &
            $\first{3.3}$ & $\first{80.2}$ & $\second{1.9}$ & $\second{0.029}$ \\
        \hline
        \bodiffo &
            ${6.9^{\pm{0.5}}}$ & ${54.0^{\pm{3.7}}}$ & $\first{1.4^{\pm{0.1}}}$ & $\first{0.021^{\pm{0.004}}}$ \\
        \egoalloho & 
            ${4.0^{\pm{0.2}}}$ & ${74.4^{\pm{2.9}}}$ & ${2.5^{\pm{0.1}}}$ & ${0.049^{\pm{0.003}}}$ \\
        \hline
        {\methodname} \textbf{(Ours)}   & 
            $\second{3.5^{\pm{0.1}}}$ & $\second{79.5^{\pm{2.1}}}$ & ${2.3^{\pm{0.1}}}$ & ${0.037^{\pm{0.002}}}$ \\
        \specialrule{.1em}{.05em}{.05em} 
    \end{tabular}
    
    \begin{tabular}{lcccc}
        \specialrule{.1em}{.05em}{.05em} 
        \multicolumn{5}{c}{\textbf{OMOMO}}\\
        \cmidrule{2-5}
                \bf Method & Min. D.$\downarrow$ & $C_{\%}\uparrow$ & $PD\downarrow$ & $P_{sc}\downarrow$ \\
         \cmidrule{2-5}
        Data             &
            $\first{1.3}$ & $\first{91.4}$ & $\first{0.9}$ & $\first{0.006}$ \\
        \hline
        \bodiffo &
            ${4.3^{\pm{0.7}}}$ & ${73.9^{\pm{5.1}}}$ & $\second{1.1^{\pm{0.2}}}$ & ${0.019^{\pm{0.004}}}$ \\
        \egoalloho &
            ${2.5^{\pm{0.2}}}$ & ${83.3^{\pm{1.8}}}$ & ${1.3^{\pm{0.1}}}$ & ${0.022^{\pm{0.002}}}$ \\
        \hline
        {\methodname} \textbf{(Ours)}   & 
            $\second{2.0^{\pm{0.1}}}$ & $\second{86.1^{\pm{1.7}}}$ & $\second{1.1^{\pm{0.1}}}$ & $\second{0.016^{\pm{0.001}}}$ \\
        \specialrule{.1em}{.05em}{.05em} 
    \end{tabular}

    \label{tab:sup_phys}
\end{table}
\renewcommand{\arraystretch}{1.0}

\renewcommand{\arraystretch}{1.2}
\begin{table}[t!]
    \setlength{\tabcolsep}{4pt}
    \centering
    \caption{\textbf{Evaluation of {\methodname} with additional input modalities}. We observe that providing {\methodname} with contact information provides the biggest quality improvement among all three modalities.}
    
    \begin{tabular}{lccccc}
    \specialrule{.1em}{.05em}{.05em} 
        \multicolumn{6}{c}{\textbf{BEHAVE}} \\
        \multirow{2}{*}{\bf Mode} & 
            \multicolumn{2}{c}{\bf Human} & & 
            \multicolumn{2}{c}{\bf Object}  \\
        \cmidrule{2-3} \cmidrule{5-6}
            & MPJPE$\downarrow$ & MPJVE$\downarrow$ & &
            $E_{v2v}\downarrow$ & $E_{c}\downarrow$ \\       
        \cmidrule{2-3} \cmidrule{5-6}
        {\methodname} & 
            ${6.82^{\pm0.08}}$ & ${7.50^{\pm0.11}}$ & &
            ${33.46^{\pm0.50}}$ & ${20.13^{\pm0.26}}$ \\
        \hline
        {\methodname} w. $\h$ &
            - & - & &
            ${32.79^{\pm0.65}}$ & ${19.41^{\pm0.31}}$ \\
        {\methodname} w. $\obj$ &
            ${6.69^{\pm0.08}}$ & ${7.37^{\pm0.10}}$ & & 
            - & - \\
        {\methodname} w. $\inter$ &
            ${6.60^{\pm0.09}}$ & ${7.22^{\pm0.12}}$ & & 
            ${32.03^{\pm0.72}}$ & ${18.59^{\pm0.28}}$ \\
    \specialrule{.1em}{.05em}{.05em} 
    \end{tabular}

    \begin{tabular}{lccccc}
    \specialrule{.1em}{.05em}{.05em} 
        \multicolumn{6}{c}{\textbf{OMOMO}} \\
        \multirow{2}{*}{\bf Mode} & 
            \multicolumn{2}{c}{\bf Human} & & 
            \multicolumn{2}{c}{\bf Object}  \\
        \cmidrule{2-3} \cmidrule{5-6}
            & MPJPE$\downarrow$ & MPJVE$\downarrow$ & &
            $E_{v2v}\downarrow$ & $E_{c}\downarrow$ \\       
        \cmidrule{2-3} \cmidrule{5-6}
        {\methodname} & 
            ${6.01^{\pm0.08}}$ & ${6.07^{\pm0.09}}$ & &
            ${26.52^{\pm1.06}}$ & ${15.23^{\pm0.28}}$ \\
        \hline
        {\methodname} w. $\h$ &
            - & - & &
            ${26.26^{\pm1.05}}$ & ${14.64^{\pm0.026}}$ \\
        {\methodname} w. $\obj$ &
            ${5.91^{\pm0.07}}$ & ${6.05^{\pm0.09}}$ & & 
            - & - \\
        {\methodname} w. $\inter$ &
            ${5.81^{\pm0.08}}$ & ${5.78^{\pm0.10}}$ & & 
            ${26.41^{\pm0.98}}$ & ${14.95^{\pm0.27}}$ \\
    \specialrule{.1em}{.05em}{.05em} 
    \end{tabular}
    
    \label{tab:sup_contacts}
\end{table}
\renewcommand{\arraystretch}{1.0}

\renewcommand{\arraystretch}{1.1}
\begin{table}[t!]
    \setlength{\tabcolsep}{4pt}
    \centering
    \caption{\textbf{Evaluation of {\methodname} with noise simulation}. We demonstrate the robustness of {\methodname} to intermittent hand tracking by randomly dropping a percentage of the input. The model maintains stable performance even with significant missing hand tracking data, confirming its resilience to sensor noise.}
    \label{tab:sup_noise}
    
    \begin{tabular}{lccccc}
    \specialrule{.1em}{.05em}{.05em} 
        \multicolumn{6}{c}{\textbf{OMOMO}} \\
            \multirow{2}{*}{\bf \%} & 
            \multicolumn{2}{c}{\bf Human} & & 
            \multicolumn{2}{c}{\bf Object}  \\
        \cmidrule{2-3} \cmidrule{5-6}
            & MPJPE$\downarrow$ & MPJVE$\downarrow$ & &
            $E_{v2v}\downarrow$ & $E_{c}\downarrow$ \\       
        \cmidrule{2-3} \cmidrule{5-6}
        \textit{0} & 
            ${6.0^{\pm0.1}}$ & ${6.1^{\pm0.1}}$ & &
            ${26.5^{\pm1.1}}$ & ${15.2^{\pm0.3}}$ \\
        \hline
        \textit{25} & 
            ${6.0^{\pm0.1}}$ & ${6.2^{\pm0.2}}$ & &
            ${26.6^{\pm1.1}}$ & ${15.4^{\pm0.3}}$ \\
        \textit{50} &
            ${6.1^{\pm0.1}}$ & ${6.3^{\pm0.2}}$ & &
            ${26.8^{\pm1.3}}$ & ${15.5^{\pm0.4}}$ \\
        \textit{75} &
            ${6.4^{\pm0.2}}$ & ${6.8^{\pm0.3}}$ & &
            ${27.3^{\pm1.6}}$ & ${16.5^{\pm0.8}}$ \\
        \textit{90} &
            ${7.7^{\pm0.5}}$ & ${8.2^{\pm0.5}}$ & &
            ${30.4^{\pm2.5}}$ & ${20.1^{\pm1.9}}$ \\
    \specialrule{.1em}{.05em}{.05em} 
    \end{tabular}
\end{table}
\renewcommand{\arraystretch}{1.0}

\renewcommand{\arraystretch}{1.1}
\begin{table}[t!]
    \setlength{\tabcolsep}{3pt}
    \centering
    \caption{\textbf{Ablation study on OMOMO}. Evaluating the impact of {\methodname} components proves the usefulness of guidance, our loss formulation, usage of three modalities, head-centric coord. system, and training with AMASS data.}
    
    \begin{tabular}{lccccc}
    \specialrule{.1em}{.05em}{.05em} 
        \multicolumn{6}{c}{\textbf{OMOMO}} \\
        \multirow{2}{*}{\bf Method} & 
            \multicolumn{2}{c}{\bf Human} & & 
            \multicolumn{2}{c}{\bf Object}  \\
        \cmidrule{2-3} \cmidrule{5-6}
            & MPJPE$\downarrow$ & MPJVE$\downarrow$ & &
            $E_{v2v}\downarrow$ & $E_{c}\downarrow$ \\
        \cmidrule{2-3} \cmidrule{5-6}
        {\methodname}                                 & 
            ${6.0^{\pm0.1}}$ & ${6.1^{\pm0.1}}$ & &
            ${26.5^{\pm1.1}}$ & ${15.2^{\pm0.3}}$ \\
        \hline
        NoGuide                 &
            ${6.1^{\pm0.1}}$ & ${6.1^{\pm0.1}}$ & &
            ${26.6^{\pm0.9}}$ & ${15.3^{\pm0.3}}$ \\
        Inpaint w/o smooth      &
            ${6.1^{\pm0.1}}$ & ${6.2^{\pm0.1}}$ & &
            ${26.6^{\pm1.1}}$ & ${15.4^{\pm0.3}}$ \\
        ($\h$, $\obj$)          &
            ${7.3^{\pm0.1}}$ & ${7.6^{\pm0.1}}$ & &
            ${26.9^{\pm0.3}}$ & ${15.7^{\pm0.2}}$ \\
        NoAMASS                 &
            ${7.4^{\pm0.1}}$ & ${7.7^{\pm0.1}}$ & &
            ${27.7^{\pm0.7}}$ & ${16.7^{\pm0.2}}$ \\
    \specialrule{.1em}{.05em}{.05em} 
    \end{tabular}

    \label{tab:sup_abl_omomo}
\end{table}
\renewcommand{\arraystretch}{1.0}

\subsubsection{Two-stage AMASS training.}
{\methodname} uniquely trains jointly on motion-only (AMASS) and HOI data in a single model.
To isolate the contribution of this formulation, we train an identical model in two stages ($\mathsf{A\text{-}FT}$: AMASS pre-train $\rightarrow$ OMOMO+BEHAVE fine-tune; Tab.~\ref{tab:sup_amass_ft}); it degrades substantially, especially on AMASS, confirming the gain comes from the tri-variate formulation, not data access.

\subsubsection{Physical plausibility.}
We evaluate physical realism in Tab.~\ref{tab:sup_phys} using established metrics: min. H-O distance (Min. D.~$[cm]$), contact \% and pen. score ($C_{\%}, P_{sc}$~$[cm]$, CHOIS~\cite{li2024controllable}), pen. depth ($PD$~$[cm]$, DiffH\textsubscript{2}O~\cite{christen2024diffh2o}).
{\methodname}'s results are close to GT, outperforming baselines that produce floaters and penetrations.
\eg, {\bodiffo}'s slightly better $P_{sc}$ stems from many floaters (only $54\%$ $C_{\%}$), while {\methodname} balances contact and penetration with $79.5\%$ $C_{\%}$ at $0.037\,cm$ $P_{sc}$.

\subsubsection{Evaluating the model with contact conditioning.}
Following the evaluation of {\methodname} with sparse $\h$ or $\obj$ tracking, we test the model's performance with $\inter$ data provided as additional conditioning.
We report the results in Tab.~\ref{tab:sup_contacts}.
Providing contact information allows for the greatest performance improvement, compared to other modalities.
This highlights that contact information plays an essential role in modeling human-object interactions.

\subsubsection{Qualitative results.}
\begin{figure*}[t!]
    \centering
    \includegraphics[trim=0cm 0cm 0cm 0cm,clip,width=0.95\linewidth]{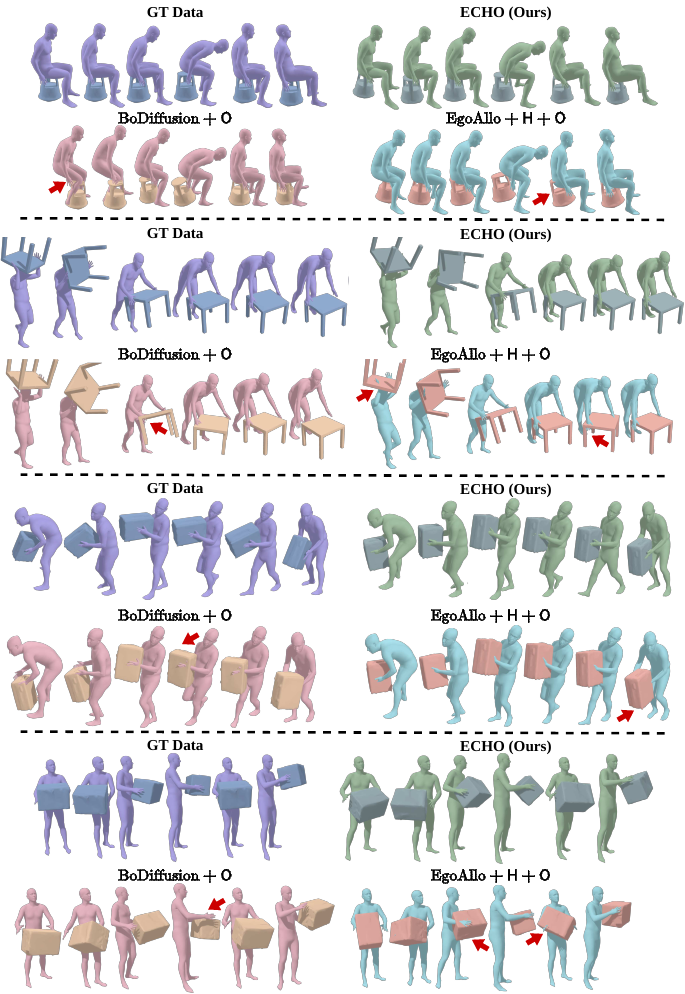}
    \vspace{-8pt}
    \caption{\textbf{Qualitative results of {\methodname}}. Our method accurately reconstructs human-object interactions across diverse scenarios. In contrast, competing methods often fail to capture correct contact dynamics, leading to artifacts such as object penetration or floating. For dynamic visualizations, please refer to the supplementary video.}
    \label{fig:sup_results}
\end{figure*}

More qualitative results of {\methodname} and comparison with baselines are provided in Fig.~\ref{fig:sup_results}.

\subsubsection{Noise simulation and ablation on OMOMO.}
We complement the evaluation of {\methodname} with noise simulation in the main paper with results on OMOMO~\cite{li2023object} in Tab.~\ref{tab:sup_noise}. 
{\methodname} demonstrates robustness to noise, maintaining stable performance even with significant missing hand tracking data.
Furthermore, we provide ablation results on OMOMO in Tab.~\ref{tab:sup_abl_omomo}.
We observe similar trends to the ablation results on BEHAVE~\cite{bhatnagar2022behave}.

\section{Implementation details}
\label{sec:a_implementation}
\subsubsection{Losses.}
The objective function used to train our network is the weighted combination of the following losses:
\begin{equation}
\begin{aligned}
    L^{\h}_{n}  &= \|\hpose - \predhpose \|_2 \\
    L^{\obj}_{n}  &= \|\objglobal - \predobjglobal \|_2 \\
    L^{\inter}_{n}  &= \|\intervec - \predintervec \|_2 \\
    L^{\obj}_{s}  &= \|\predobjvelocity - \tau_{vel} \|_2 +
        \|\predobjangvelocity - \tau_{ang} \|_2 \\
    L^{\h}_{j}  &= \|\hjoints - \predhjoints \|_2 \\
    L^{\h}_{s}  &= \|\intervec^{feet} * \predhvelocity^{feet}\|_2
\end{aligned}
\end{equation}
where $\predobjvelocity$ is the velocity of predicted object keypoints, $\predobjangvelocity$ is the angular velocity of predicted object keypoints, $\predhjoints$ are predicted human joints inferred from SMPL, $\predhvelocity^{feet}$ is the velocity of predicted feet joints, and $\intervec^{feet}$ is the ground-truth binary contact labels for feet. The resulting loss function is:
\begin{equation}
    \begin{aligned}
    L_{{\methodname}} =
        & \lambda^{\h}_{n} L^{\h}_{n} + 
        \lambda^{\obj}_{n} L^{\obj}_{n} + 
        \lambda^{\inter}_{n} L^{\inter}_{n} + \\
        & \lambda^{\obj}_{s} L^{\obj}_{s} + 
        \lambda^{\h}_{j} L^{\h}_{j} + 
        \lambda^{\h}_{s} L^{\h}_{s}
    \end{aligned}
\end{equation}
with weighting coefficients set to: $\lambda^{\h}_{n}=\lambda^{\obj}_{n}=5.0, \lambda^{\inter}_{n}=1, \lambda^{\obj}_{s}=\lambda^{\h}_{j}=\lambda^{\h}_{s}=0.01$.

\subsubsection{Inference guidance.}
We adopt a classifier-based guidance~\cite{dhariwal2021diffusion} approach at inference time. 
We formulate the guidance loss to ensure that the predicted human and object meshes align with the predicted contact.
The function therefore includes two terms: one for human-object contact and one for foot-floor contact.
The human-object term forces the contacts inferred from predicted human and object meshes to align with the contacts diffused by the network:
\begin{equation}
    \mathcal{F}_\text{HOI}(\hat{\h}, \hat{\obj}, \hat{\inter}) = 
        \frac{1}{|\interpoints|} \left< \vec{\hat{d}}, \predintervec^{\text{HOI}} \right> 
\end{equation}
where $\predintervec^{\text{HOI}}$ is a vector of predicted human-object contacts, $\vec{\hat{d}} \in \mathbb{R}^{64}$ is a vector of distances between the subset $\widehat{\interpoints}$ of predicted human mesh vertices $\predhtemplatev$ and object mesh vertices $\predobjtemplatev$:
\begin{equation}
    \mathbf{\hat{d}}_j= 
        \min_{i = 1, \dots, |\predobjtemplatev|} \left\| \widehat{\interpoints}^j - \predobjtemplatev^i \right\|_2, \quad j = 1, \dots, |\widehat{\interpoints}|.
\end{equation}

For human-floor interaction, we penalize excessive foot skating based on the dynamics of predicted contacts. Similarly to \cite{yi2025egoallo}, we define:
\begin{equation}
    \mathcal{F}_\text{skate}(\hat{\h}, \hat{\obj}, \hat{\inter}) =
        \sum_{t, j}{\left\| \frac{1}{2}
            \left(
                \left(\predintervec^{\text{Env}}\right)^t_j +
                \left(\predintervec^{\text{Env}}\right)^{t-1}_j
            \right)
            \left( 
                \left(\predhjoints\right)^t_j +
                \left(\predhjoints\right)^{t-1}_j
            \right) 
        \right\|_2}
\end{equation}
where $\predintervec^{\text{Env}}$ is a vector of predicted human-environment contacts, $\predhjoints$ are predicted human body joints of the SMPL model, the summation is done over time $t$ within window length $W$ and selected joints $j$ (i.e., ankles and toes).

The final loss function for the guidance is a weighted sum of the above terms, with $\lambda_\text{HOI}=150, \lambda_\text{skate}=0.25$:
\begin{equation}
    \mathcal{F}(\hat{\h}, \hat{\obj}, \hat{\inter}) = 
        \lambda_\text{HOI}\mathcal{F}_\text{HOI}(\hat{\h}, \hat{\obj}, \hat{\inter}) +
        \lambda_\text{skate}\mathcal{F}_\text{skate}(\hat{\h}, \hat{\obj}, \hat{\inter})
\end{equation}

We adopt the reconstruction guidance formulation of \cite{ho2022video}, where the predicted sample $(\hat{\h}, \hat{\obj}, \hat{\inter}) = \methodnetwork(\h, \obj, \inter; \thuman, \tobj, \tinter, \condobj, \condego)$ is directly modified on each denoising step. 
The reconstruction guidance with scale $\lambda=0.1$ is thus formulated as:
\begin{equation}
    (\hat{\h}, \hat{\obj}, \hat{\inter}) \coloneq 
        (\hat{\h}, \hat{\obj}, \hat{\inter}) - 
        \lambda \nabla_{\h^{\set{T}_\h}, \obj^{\set{T}_\obj}, \inter^{\set{T}_\inter}} 
            \mathcal{F}(\hat{\h}, \hat{\obj}, \hat{\inter}).
\end{equation}

\subsubsection{Smooth inpainting.}
To ensure smooth transitions during inference on long sequences, we introduce \textit{smooth inpainting}. 
This method extends standard inpainting inference~\cite{guzov-jiang2025hmd2} by blending new predictions with past ones in the overlapping window region, rather than discarding them. 
The blending is performed at each diffusion step according to the following formula:
\begin{equation}
    \begin{aligned}
        \hat{\h}^{\set{T}_\h}_\mathcal{W} &\coloneq \alpha\hat{\h}^{\set{T}_\h}_\mathcal{W} + (1-\alpha)\hat{\h}_{\mathcal{W}-1}, \\
        \hat{\obj}^{\set{T}_\obj}_\mathcal{W} &\coloneq \alpha\hat{\obj}^{\set{T}_\obj}_\mathcal{W} + (1-\alpha)\hat{\obj}_{\mathcal{W}-1}, \\
        \hat{\inter}^{\set{T}_\inter}_\mathcal{W} &\coloneq \alpha\hat{\inter}^{\set{T}_\inter}_\mathcal{W} + (1-\alpha)\hat{\inter}_{\mathcal{W}-1},
    \end{aligned}
\end{equation}
where $\hat{\h}^{\set{T}_\h}_\mathcal{W}, \hat{\obj}^{\set{T}_\obj}_\mathcal{W}, \hat{\inter}^{\set{T}_\inter}_\mathcal{W}$ are the predictions for the set of denoising steps $\set{T}_\h, \set{T}_\obj, \set{T}_\inter$ for the current window $\mathcal{W}$, $\hat{\h}_{\mathcal{W}-1}, \hat{\obj}_{\mathcal{W}-1}, \hat{\inter}_{\mathcal{W}-1}$ are the predictions for the previous window $\mathcal{W}-1$, and $\alpha=0.4$ is a blending factor.

\section{Metrics}
\label{sec:a_metrics}
\subsubsection{Evaluating human prediction.}
MPJPE measures the average $L_2$ distance between predicted $\predhjoints$ and ground-truth body joints $\hjoints$:
\begin{equation}
    \text{MPJPE}(\hjoints, \predhjoints) = \frac{1}{|\hjoints|}\sum\limits_{i \in |\hjoints|}{\|\hjoints^i - \predhjoints^i\|_2}
\end{equation}

MPJVE measures the average velocity error for predicted $\predhjoints$ and ground-truth body joints $\hjoints$. 
Velocity at step $i$ is computed as: $\hvelocity^i = \hjoints^i - \hjoints^{i-1}$, thus we define:
\begin{equation}
    \begin{aligned}
        & \text{MPJVE}(\hjoints, \predhjoints) = 
         \frac{1}{|\hjoints| - 1}
            \sum\limits_{i \in \{1..|\hjoints|\}}
            {\|\hvelocity^i - \predhvelocity^i\|_2}
    \end{aligned}
\end{equation}

Foot Contact (FC) measures the fraction of frames with any of 4 feet joints (ankle and foot for both legs) located closer to the ground than a pre-defined threshold (10 and 5 cm, respectively).

\subsubsection{Evaluating object prediction.}
$E_{v2v}$ measures the average $L_2$ distance between the positions of the predicted object vertices and the ground-truth ones:
\begin{equation}
    E_{v2v}(\objtemplatev, \predobjtemplatev) = \frac{1}{|\objtemplatev|}\sum\limits_{i \in \{0..|\objtemplatev|\}}{\|\objtemplatev^i - \predobjtemplatev^i\|_2}
\end{equation}

$E_{c}$ measures the average $L_2$ distance between the position of the predicted object center and the ground-truth one:
\begin{equation}
    \begin{aligned}
        & E_{c}(\objtemplatev, \predobjtemplatev) =  
         \left\Vert{
            \frac{1}{|\objtemplatev|}
                \sum\limits_{i \in \{0..|\objtemplatev|\}}{\objtemplatev^i} - 
            \frac{1}{|\predobjtemplatev|}
                \sum\limits_{i \in \{0..|\predobjtemplatev|\}}\predobjtemplatev^i
        }\right\Vert_2
    \end{aligned}
\end{equation}

Rotation Difference (Rot. Diff.) measures the average angular difference between predicted global rotation for the object and the ground-truth one.

The contact accuracy metric $\text{Acc}_{\inter}$ is defined as the accuracy between the ground-truth binary contact vector and a binary contact vector inferred by thresholding a vector of distances $\vec{\hat{d}}$ between predicted human and object meshes.

\end{document}